# Dynamic neighbourhood optimisation for task allocation using multi-agent learning

NIALL CREECH, Kings College London, UK

NATALIA CRIADO PACHECO, Kings College London, UK

SIMON MILES, Kings College London, UK

In large-scale systems there are fundamental challenges when centralised techniques are used for task allocation. The number of interactions is limited by resource constraints such as on computation, storage, and network communication. We can increase scalability by implementing the system as a distributed task-allocation system, sharing tasks across many agents. However, this also increases the resource cost of communications and synchronisation, and is difficult to scale.

In this paper we present four algorithms to solve these problems. The combination of these algorithms enable each agent to improve their task allocation strategy through reinforcement learning, while changing how much they explore the system in response to how optimal they believe their current strategy is, given their past experience. We focus on distributed agent systems where the agents' behaviours are constrained by resource usage limits, limiting agents to local, rather than system-wide knowledge. We evaluate these algorithms in a simulated environment where agents are given a task composed of multiple subtasks that must be allocated to other agents with differing capabilities, to then carry out those tasks. We also simulate real-life system effects such as networking instability. Our solution is shown to solve the task allocation problem to 6.7% of the theoretical optimal within the system configurations considered. It provides 5× better performance recovery over no-knowledge retention approaches when system connectivity is impacted, and is tested against systems up to 100 agents with less than a 9% impact on the algorithms' performance.

## 1 INTRODUCTION

In a *distributed task-allocation system (*DTAS*)* there are interactions between many independent agents. These systems are increasingly seen in a wide range of real world applications such as wireless sensor networks (WSN) [5, 7, 36, 50], robotics [12, 46], and distributed computing [38, 48]. The growing complexity and scope of these applications presents a number of challenges such as responding to change, handling failures, and optimisation. System performance must

Authors' addresses: Niall Creech, Kings College London, Department of Informatics, London, WC2B 4BG, UK, niall.creech@kcl.ac.uk; Natalia Criado Pacheco, Kings College London, Department of Informatics, London, WC2B 4BG, UK, natalia.criado_pacheco@kcl.ac.uk; Simon Miles, Kings College London, Department of Informatics, London, WC2B 4BG, UK, simon.miles@kcl.ac.uk.

also be scalable with growth in the number of agents, being able to perform tasks given constraints in computational or storage resources. The challenges summarised below are shared across many diverse subject areas, meaning relevant and practical solutions become more generally applicable.

- *task allocation*, how best to allocate tasks amongst agents in the system. An agent may have a goal that comprises of a composite task that requires the completion of a number of sub-tasks by other agents [70].
- *resource management*, allocating and optimising the use of resources to complete a task. For example, managing energy usage while performing a function within a physical environment [29, 60, 96].
- *dynamic networking*, agent discovery and communication adaptability. Agents must be able to communicate with each other while connections are lost and created [6].
- *self-organisation*, autonomously forming structures to complete a goal. Solutions with rigid architectures are often non-applicable to dynamic systems with many unknowns as designs would be too complex. To improve agents adaptability in these situations, self-organising solutions can be used. [1, 26, 27, 34, 47].

Formally designed agents can perform set tasks given a well-understood system. However, it is often not feasible to design algorithms that can predict the large variety of failures or changes that may occur in large-scale, real-world operating environments. In addition, as the systems become more complex there is an exponential growth in agents *state-action space* size. This space represents the set of combinations of states they can be in, alongside the actions they may take in those states. Knowing this space before deploying the agents is often unrealistic, as is understanding which algorithms will perform optimally. Introducing a centralised source of continually updated information on the environment and other agents can increase the knowledge available to an agent about their state-action space, allowing for better optimisation. Approaches like this such as the use of *orchestrating agents*, agents that specialise in coordinating other agents in the system, are used within distributed software architectures [39, 41, 49, 66] and robotics [4, 20]. However, in extending this method through clustering and consensus techniques to increase fault-tolerance, a central point of fragility is created. As other agents' interactions and communications are channelled through these centralised agents, congestion and bandwidth saturation problems also grow.

Distributed agent systems with learning enhancements such as *multi-agent reinforcement learning (MARL)* can provide the same functionality but distributed across agents, removing the focal points for orchestration and mitigating congestion issues while still providing the knowledge sharing and action coordination that allow agents to optimise state-action space. With an increasing number of interacting agents though we see an exponential increase in the amount of communications within the system, eventually saturating bandwidth and exhausting computational resources. There is also an *expectation of stability*, that the solution to the agents optimisation remains relatively stable with a gradual reduction in the need for exploration of state-action space over time. In dynamic systems this often does not hold. MARL techniques also do not take account of the inherent risks involved in taking different types of actions, leading to catastrophic effects in areas such as robotics where some actions may risk severe physical damage, or in financial systems where large losses might be incurred [33, 40, 57, 87].

The overall problem can be summarised as how to provide for efficient task allocation in a dynamic multi-agent system while ensuring scalability as the number of tasks increases and the availability of agents changes. The solution presented uses a number of algorithms in combination, allowing an agent to determine the capability of other known agents to perform tasks, allocating these tasks, and carrying out other actions based on its current knowledge and the need to explore agent capability space. The algorithms introduced are:

- the *agent task allocation with risk-impact awareness (*`ATA-RIA`*)* algorithm allows each agent to choose a subset of other agents in the system based on how much it predicts those agents will help complete the sub-tasks of their overall composite task. They can learn the best task allocation strategy for these agents, but can also change which agents compose the group to improve performance.
- the *reward trends for action-risks probabilities (*`RT-ARP`*)* algorithm gives agents the ability to transform their exploration strategies given the trends in the rewards obtained over time. Using this algorithm, agents can increase the likelihood of them taking actions that risk larger changes to their task allocation strategy, depending on their historical performance.
- the *state-action space knowledge-retention (*`SAS-KR`*)* algorithm intelligently manages the resources used by agents to maintain the information they have learned about state-action space and the effects of their actions.
- the *neighbourhood update (*`N-Prune`*)* algorithm selectively removes agents from the group considered for task allocation by an agent, constraining resource usage. This selection is based on not only how much an agent predicts the other agents will contribute to its composite task, but also how much uncertainty it has about that prediction, so complimenting the `ATA-RIA` algorithms' behaviour.

We test the effectiveness of these algorithms through evaluation of their performance in a series of simulated multi-agent systems.

Section 2 covers the related research in the areas of MARL and multi-agent systems. In-depth analysis of the problem domain and motivation is looked at in Section 3, with the proposed solution and algorithm definitions in Sections 4 and 5. We cover evaluation of the algorithms' performance in system simulations in Section 6. Finally we discuss conclusions and future research in Section 8.

## 2 RELATED WORK

To provide some context for the work to follow we look at some relevant research in multi-agent reinforcement learning (MARL). Although there are other useful strategies, such as auction-based systems, and particle swarm optimisation techniques, these also have specific challenges. Auction-based systems carry increasing orchestration cost as the number of agents involved increases, which impacts the scalability of related solutions. They also suffer from performance loss when the system is dynamic as agent communication is disrupted. Swarm approaches can be effective under dynamic conditions but are also prone to optimising on local-optima [76]. As we look for an approach that can handle scaling, and dynamic systems, we focus here on MARL. In particular, we look at ways of allocating rewards to drive behaviours, and how allocation effects both the exploration of state space, and coordination between agents.

Multi-agent reinforcement learning (MARL) [16, 17, 85] applies reinforcement learning techniques to multiple agents sharing a common environment. Each senses the environment and takes actions that cause a transition of the environment state to a new state, resulting in feedback in the form of the reward signal. There are a number of issues that can limit the applicability of MARL techniques which we discuss next.

### 2.1 Challenges of high-dimensionality systems

As the number of agents in these systems increase, there is a corresponding exponential increase in the possible communications and actions an agent may take with respect to other agents in the system. This increases the state-action space size, limiting the scale of systems that standard learning algorithms can be applied to. There has been much work in making these large state-spaces tractable for computation. Through *aggregation* or *abstraction*, the number of states can be reduced through combining similar ones into a single state in the learning model [3, 88]. This simplifies the model, but sacrifices information about the merged states. Additionally, it can be difficult to qualify which states are similar enough to be abstracted, and the effect of doing so on the agents' performance can be unpredictable in more complex multi-agent systems.

With *state-space generation* and *adaptation* algorithms, we can have the algorithm generate its own initial state-space representation [65], and adapt this representation throughout its lifetime [28, 35]. This approach reduces the state-space down to those states relevant to an agents' learning function, while ignoring the others. Our SAS-KR algorithm, discussed in Section 4.3, develops on this approach, allowing an agent to not only generate and adapt its known states through its lifetime, but also to forget state information that is judged to be less valuable to its success. In addition, the N-Prune algorithm reduces an agents' state-space by restricting the number of other agents it can observe at one time, avoiding many of the challenges of large state-spaces.

## 2.2 Exploration in large, non-stationary environments

Finding the right balance of exploration, so that agents' can discover the optimal actions in expansive state spaces, and exploitation, so that they can successfully complete tasks, is difficult [9, 61]. Using *undirected methods* [84], where exploration is effectively at random, is not feasible in large state-spaces where the sparseness of action sampling slows learning. For this reason we focus on *directed methods* where knowledge can be used to make algorithms more selective in searching state-space.

The exploration/exploitation challenge increases in difficulty with the dynamism of the policies and actions of other agents. In stationary environments, there is often an initial highly explorative stage, commonly using $\epsilon$-*greedy* action selection [90], which then switches off in favour of a continual exploitation stage once the algorithms performance is deemed to be acceptable. This may also take the form of a decay factor, where the degree of exploration decreases gradually over time such as in standard *Boltzmann exploration* [61]. In a non-stationary environment however, the tasks and their distribution may change. Agents may affect the environment and be effected by the behaviours of other agents. A time or performance based switch to exploitation risks a reduction in algorithm performance as the most optimal actions continue to change over time, but the agents' probabilities of choosing actions remain static. *Adaptive exploration* techniques [9] are designed for this non-stationary situation, varying exploration and exploitation throughout the system lifetime [80]. Other algorithms increase the exploration of infrequently sampled actions. Examples of this are count-based approaches [94] extending Boltzmann exploration with a state-action visitation factor [94]. Successor representations [23] have also been used as the state-action sampling metric to incentivise exploration [54].

While these methods can work in non-stationary environments where the degree of change is relatively constant, often the rate of change can accelerate or decelerate, or be relatively static in some areas of the system and highly dynamic in others. For example, in an ocean-based environment, currents might be volatile and rapidly changing in one part of the environment, but be stable with calm seas in another. To work in those environments, we use a variation of these approaches that utilises state-action space sampling history as well as past rewards history to guide exploration for our RT-ARP algorithm, discussed in Section 4.2.

## 2.3 No-regret exploration and intrinsic motivation

These exploration strategies use the principle of *optimism in the face of uncertainty*, the assumption that less well-known state-actions are worth exploring [68]. The use of *no-regret* to optimise reinforcement learning algorithms is well established [44] , with additional work applying this to exploration strategies [63, 83]. Agents can also be given different *intrinsic motivations*, underlying goals that generate rewards in addition to immediate task-completion benefits. Methods such as knowledge acquisition [67] or Bayesian curiosity [14] can then be used to drive exploration behaviours. Short-term and long-term intrinsic rewards can be combined to encourage local, and deeper system exploration respectively [15]. We look to improve on this work by adapting how optimistically an agent explores, not only based on uncertainty, but how optimally it believes it is exploiting the system given its past history.

The `RT-ARP` algorithm introduces a form of regret-minimisation exploration based on a function of the rewards over long and short-term timescales. We detail this work in Section 5.5.4 in which we describe how our *impact transformation function* is used by agents to predict the risk of taking more disruptive actions, and exploring more aggressively. This also provides a degree of risk-based intrinsic motivation, where agents are encouraged to explore more when short and long-term success are unequal, and focus more on sampling-based Boltzmann exploration when they are comparable.

### 2.4 The stability-plasticity dilemma in continual learning

One of the challenges in non-stationary reinforcement learning is how much knowledge should an agent preserve about its past experiences compared to adapting to more recent ones. This *stability-plasticity dilemma* affects how well agents complete new tasks they have previously seen [71, 77]. In the worst case it can result in *catastrophic inference* [64], where tasks an agent has completed in the past are treated as completely unknown when seen again in the future. The optimal balance of stability and plasticity is dependent on the proportion of tasks an agent sees in the future that will be similar to ones it has seen in the past. This is often achieved through *experience replay*, ensuring that past events are reapplied in the current learning context so as to not be completely overwritten by updates due to an agents' present actions [42, 73, 75] or localised learning updates to reduce overwriting past learned action probabilities [52, 97]. There are difficulties however in selecting which past experiences are relevant in the present and should be reapplied. Successful or rewarding past actions may not be useful in an agents' current context given the non-stationary nature of the environment.

The `RT-ARP` algorithm helps to address the stability-plasticity challenge by measuring learning success over a range of short to longer term time-scales, then adapting the speed of learning based on the comparison of task rewards over these periods, discussed in Section 5.5. The effect of this is that plasticity is increased. Behaviours are more strongly overwritten when the agents current policy is performing well in the short-term, but poorly over the longer term. As shorter and longer-term rewards become comparable, plasticity is decreased and learned values become more stable.

### 2.5 Coordination in agent-based systems

In general, coordination in multi-agent systems increases the optimality of solutions found, but at the cost of increased overhead which limits scalability. Agents in MARL systems can range from being fully cooperative to fully competitive. In cooperative systems the agents all share a common reward function and try to maximise that shared value function. Dedicated algorithms often rely on static, deterministic, or on exact knowledge of other agents' states and actions. Coordination and maximisation of joint-action states results in high dimensionality due to the inclusion of the actions of other agents in calculations. To avoid this overhead, we can utilise the sparseness of the interactions in large multi-agent systems to reduce the coupling between agents by having them work independently and only collecting information about other agents when required. For example, by learning the states where some degree of coordination is needed [24, 25, 62].

Similarly, when approaching tasks that can be decomposed and allocated amongst a group of agents in a multi-agent system, we can use decomposed reward signals to induce some degree of coordination amongst localised agents that share those subtasks [81]. In a non-stationary environment, the value of those tasks to the allocating agent, and the capability of those agents completing subtasks, can change, discussed in Section 3.3.1.

## 2.6 Summary of key challenges

This past research highlights some of the key challenges that we look to tackle in our work,

(1) in large or complex systems the correct policies for agents' behaviour are not known at system initialisation, and may be constantly changing due to system dynamics.

(2) since systems may be dynamic, the optimal solution may be constantly changing.

(3) for a scalable system, system-wide knowledge is not feasible to maintain or to compute with.

(4) agents have physical constraints on compute and memory in real situations that limit their maximum resource usage.

To do this we need to develop the abilities for agents to,

(1) learn to make the best decisions given their current state.

(2) adapt how they explore state-space depending on how successful they are in task-allocation currently.

(3) make decisions based only on a localised or otherwise partial-view of the system.

(4) maintain their resource usage within set limits.

The four algorithms we present in the following sections are designed to tackle these issues and combine to form a scalable, resilient, and adaptive multi-agent task allocation solution.

## 3 TASK ALLOCATION IN MULTI-AGENT SYSTEMS

In the following sections we introduce the multi-agent system problem and model the system.

### 3.1 Distributed Task Allocation System

Informally we define a distributed task allocation system as a multi-agent system where a set of agents work together to perform a set of *composite tasks*. These composite tasks are formed by *atomic tasks* that can be executed by individual agents. Each agent has some capabilities to perform atomic tasks and is also able to coordinate and oversee the execution of a set of composite tasks. Each agent also has constraints on memory and communication, limiting the number of agents it can interact with and maintain information on. This in turn constrains the size of the set of agents it can learn to allocate tasks to, and the amount of knowledge it can retain on the systems' agents overall.

*Definition 3.1 (Distributed Task Allocation System).* A distributed task-allocation system (DTAS) is defined by a tuple $\langle AT, CT, G \rangle$ where:

- $AT$ is the set of atomic tasks $at$ (or tasks for short), where each task can be performed by a single agent;
- $CT$ is the set of composite tasks $ct$, where each composite task is formed by a set of atomic tasks;
- $G$ is the set of agents, where each agent $g$ is defined by a tuple $\langle id, c, r, \delta_n, \delta_k \rangle$, where:
  - $id$ is a unique identifier for the agent;
  - $c \subseteq AP$ is the agent capabilities; i.e., the atomic task types that the agent can perform;
  - $r \subseteq CP$ is the agent responsibilities; i.e., the composite task types that the agent can oversee;
  - $\delta_n, \delta_k \in \mathbb{N}$, are the resource constraints of the agent, namely the communication and memory constraints (i.e., how many other agents a given agent can communicate with and know about).

Atomic tasks are of one of the *atomic task types ap* in the system, with *composite task types cp* defined by the type of its elements. For a set of atomic tasks, *AT*, and a set of composite tasks, *CT*, we define $type_a \colon AT \to AP$ and $type_c \colon CT \to 2^{AP}$ as the mappings of atomic and composite tasks to their respective task types, where $type_c(\{at_1, .., at_n\}) = \{type_a(at_1), .., type_a(at_n)\}$.

Given an agent $g$, we denote by $c(g), r(g), \delta_n(g), \delta_k(g)$ the capabilities, responsibilities, communication, and memory constraints of that agent, respectively. These communication constraints limit the number of other agents that an agent can interact with at any one time, its *neighbourhood*, while memory constraints limit the amount of information it can have about other agents in the system as a whole, its *knowledge*. Note that for all atomic tasks in the system there is at least one agent capable of performing it. Similarly, for all composite tasks in the system there is at least one agent responsible for overseeing it.

### 3.2 System Dynamics

Composite tasks arrive in the system with constant or slowly varying frequency distribution. The DTAS is capable of processing these tasks in the following way:

(1) a request to perform composite task of a defined composite task type arrives in the system.
(2) the composite task is allocated to an agent that can be responsible for tasks of that type.
(3) the agent decomposes the composite task into atomic tasks.
(4) the agent allocates these atomic tasks to other agents.
(5) once all the atomic tasks have been completed the composite task is complete.

To be able to allocate atomic tasks, agents need to not only be aware of the other agents in the system and their capabilities to execute tasks, but also to have communication links with them. Hence, the current state of an agent is determined by the agents it knows (i.e., its knowledge) and the agents it has links with (i.e., its neighbourhood).

*Definition 3.2 (Agent State).* Given an agent $g = \langle id, c, r, \delta_n, \delta_k \rangle$, we define its state at a particular point in time as a tuple $\langle K, N \rangle$, where:

- $K \subseteq G$ is the knowledge of the agent[1].
- $N \subset K$ is the neighbourhood of the agent.

Note that $|K| \le \delta_k$ and $|N| \le \delta_n$. Given an agent $g$ we denote by $K(g), N(g)$, its knowledge and neighbourhood. Given a set of agents $G$, we denote by $G_S$ the set formed by their states.

At a given point in time the system is required to perform a set of composite tasks $R$ by a set of external agents $E$. For simplicity, we assume that only one request can be done at a given moment in time and, hence, time allows us to distinguish between different requirements to perform the same task. Therefore it acts as an identifier for each composite task, and the associated atomic tasks, allocated to the system.

A requirement to perform a composite task is allocated to a particular agent. We represent this by tuples such as $\langle ct, t, e, pg \rangle$, where $ct \in CT$, $t \in \mathbb{N}$ is the time at which the request to perform the task was created; $e \in E$ is the agent who requested the execution of the composite task and $pg \in G$ is the agent responsible for the completion of the composite task, the *parent agent*. Agents can also be allocated atomic tasks that are needed to complete a composite task, which we term *child agents*, $cg \in G$. We represent that as allocations where a set of tasks is formed by one task $\langle\{at\}, t, pg, cg\rangle$, where $cg$ is capable of performing the atomic task $at$. In general, we denote by $L$ the set of all allocations at a given point in time, containing both composite, and atomic tasks. The set is formed by tuples $\langle T, t, epg, pcg\rangle$ where $T$ is a list of atomic tasks (which can be defined as a composite task), $t \in \mathbb{N}$ is the time at which the request to perform the task was created, $epg \in (G \cup E)$ is the parent or external agent which allocated the task, and $pcg \in G$ is the parent or child agent which is allocated the task.

---

[1]For simplicity, we represent the knowledge about a particular agent by the agent identifier, but the knowledge could also include other information such as agent capabilities and qualities when performing particular actions, etc.

*Definition 3.3 (System State).* Given a DTAS we define its state as a tuple $S = \langle G_S, AL_S\rangle$ where

- $G_S$ is the set of states of all agents in the system;
- $AL_S$ is the set of task allocations in the system.

*Example 3.4 (Real-world systems).* A marine-based WSN system agents are equipped with sensors that can complete tasks to measure temperature, salinity, oxygen levels, and pH levels, so $AP = \{ap_{temp}, ap_{sal}, ap_{oxy}, ap_{ph}\}$. Each agents' capabilities may be a subset of these atomic task-types depending on which sensors they have, and whether they are functional. For instance $c_g = \{ap_{sal}, ap_{oxy}\}$, if an agent $g$ only has working sensors to measure salinity and oxygen levels. Some agents receive composite tasks from outside the system, requests for samples of combinations of these measurements, e.g. $ct = \{at_{sal}, at_{oxy}\}$. These agents then decompose these composite tasks into atomic tasks and allocate them to other agents to complete.

### 3.2.1 Actions.
The DTAS's configuration changes as a result of the actions executed by the agents and actions taken by the external agents (e.g., users) who make requests to the system to execute a set of tasks. In the following we provide the operational semantics for the different actions that can be executed in a DTAS.

- Requirement Assignment. Every time the DTAS receives a new requirement from an external agent $e$ to perform an composite task $ct$ at a given time $t$ it is randomly assigned to an agent responsible for that task:

$$\frac{REQUIREMENT(e, ct) \wedge e \in E \wedge time(t) \wedge \exists g \in G : ct \in r(g)}{\langle G_S, AL_S\rangle \rightarrow \langle G_S, AL_S \cup \{\langle ct, t, g, e\rangle\}\rangle}$$

  where $g$ is a randomly selected agent being responsible for that composite task and *time* just returns the current time of the DTAS.

- Allocation action. A agent $g$ performing an allocation action allocates an atomic task that is currently allocated to him to a single neighbourhood agent, which can then not be allocated to another agent, or re-allocated. The system state is updated accordingly:

$$\frac{ALLOC(g, at, n) \wedge g \in G \wedge at \in AT \wedge n \in N(g) \wedge \exists\langle T, t, g, a\rangle \in AL_S : at \in T}{\langle G_S, AL_S\rangle \rightarrow \langle G_S, AL_S \cup \{\langle\{at\}, t, n, g\rangle\}\rangle}$$

- Execute action. If an agent is allocated an atomic task and is capable of performing it $at \in c(g)$ then it can perform an execute action, $EXEC(g, at)$:

$$\frac{EXEC(g, at) \wedge g \in G \wedge at \in AT \wedge at \in c(g) \wedge \exists\langle T, t, g, a\rangle \in AL_S : at \in T}{\langle G_S, AL_S\rangle \rightarrow \langle G_S, AL_S'\rangle}$$

  where $AL_S' = \{\langle T, t', g, a\rangle | \langle T, t', g, a\rangle \in AL_S \wedge t' <> t\} \cup \{\langle T', t, g, a\rangle | \langle T, t, g, a\rangle \in AL_S \wedge T' = T \setminus \{at\}\}$. After executing an atomic task with a given time identifier, all tasks allocations corresponding to that identifier are reviewed so that the atomic task is removed from the list of pending tasks.

- Information action. An agent can request information on other agents in the system, from an agent in its neighbourhood, by carrying out an info action.

$$\frac{INFO(g,t,n) \land g \in G \land time(t) \land n \in N(g)}{\langle G_S, AL_S \rangle \rightarrow \langle G_S, AL_S \cup \{\langle \{info\}, t, n, g \rangle\}\rangle}$$

where $info$ is an special information atomic task that is not part of any composite task.

- Provide Information. Agents who are allocated an info action execute that action by providing information about one of their randomly selected neighbourhood agents, $u$:

$$\frac{PROVIDE\_INFO(g,n,u) \land g \in G \land n \in N(g) \land u \in K(g) \land \langle \{info\}, t, g, a \rangle \in AL_S}{\langle G_S, AL_S \rangle \rightarrow \langle G'_S, AL_S \setminus \{\langle \{info\}, t, g, a \rangle\}\rangle}$$

where $G'_S = \{\langle K(g'), N(g')\rangle | \forall g' \in (G \setminus \{n\}\} \cup \{\langle K(n) \cup u, N(n)\rangle\}$

- Remove Info: An agent $g \in G$ can remove information about an agent from its knowledge as long as that agent is not in its neighbourhood:

$$\frac{REMOVE\_INFO(g,k) \land g \in G \land k \in K(g) \land k \notin N(g)}{\langle G_S, AL_S \rangle \rightarrow \langle G'_S, AL_S \rangle}$$

where $G'_S = \{\langle K(g'), N(g')\rangle | \forall g' \in (G \setminus \{g\}\} \cup \{K(g) \setminus \{k\}, N(g)\}$

- An agent can add a known agent into its neighbourhood by taking a link action, $LINK(g,k)$:

$$\frac{LINK(g,k) \land g \in G \land k \in K(g) \land |N(g)| < \delta_n(g)}{\langle G_S, AL_S \rangle \rightarrow \langle G'_S, AL_S, t, g\rangle\})}$$

where $G'_S = \{\langle K(g'), N(g')\rangle | \forall g' \in (G \setminus \{g\}\} \cup \{\langle K(g), N(g) \cup \{k\}\rangle\}$

- Remove Link. An agent $g \in G$ can remove an agent $n$ from its neighbourhood by taking a remove link action, $REMOVE\_LINK(g,n)$:

$$\frac{REMOVE\_LINK(g,n) \land g \in G \land n \in N(g)}{\langle G_S, AL_S \rangle \rightarrow \langle G'_S, AL_S \rangle}$$

where $G'_S = \{\langle K(g'), N(g')\rangle | \forall g' \in (G \setminus \{g\}\} \cup \{K(g), N(g) \setminus \{n\}\}$

We map a given action $a$ to one of the defined action-categories above as $category(a)$. Every action will return a quality value. The quality values returned by $ALLOC$ actions will be discussed in Section 3.3.1. We set the quality value for all other actions to be zero.

*Example 3.5 (Actions).* An agent $g$ in a marine WSN with a neighbourhood $\{g_1, g_2, g_3\}$, receives a composite task $ct = \{at_{sal}, at_{oxy}\}$. Since agent $g$ has a working salinity measuring sensor, $ap_{sal} \in c_g$, it can complete the task $at_{sal}$ itself, and so performs action $EXEC(g, at_{sal})$. As it doesn't have a sensor to detect oxygen levels, it cannot complete tasks of that type, $ap_{oxy} \notin c_g$, and so it allocates this task to an agent in its neighbourhood, $g_1$, through the action $ALLOC(g, at_{oxy}, g_1)$.

### 3.2.2 Specifying groups of actions.
Given a set of actions $A$, let $actions: \mathcal{A} \times G \rightarrow \mathcal{A}$ be all the actions that can be taken by a given agent $g$. We define *target actions* of an agent, $targets: \mathcal{A} \times G \times \mathcal{G} \rightarrow \mathcal{A}$, as those actions in the set of all actions that have arguments containing an agent in a set of agents $G$

### 3.3  Task quality and the optimality of allocations

*3.3.1  Task and allocations quality.* We denote all possible *allocations* of atomic tasks to agents in the system, $\mathcal{AL} = 2^{(\mathcal{AT} \times \mathcal{G})}$. Given a set of atomic tasks, and a set of agents, there are a number of different *permutations*: $2^{\mathcal{AT}} \times 2^{\mathcal{G}} \rightarrow 2^{\mathcal{AL}}$ allocating these tasks amongst the agents where,

$$permutations(AT, G) = \{AL \in \mathcal{AL} \mid \forall (at, g) \in AL, at \in AT, g \in G\} \tag{1}$$

Given an allocation, those tasks that have been allocated to an agent $g$ are given by its *concurrent allocations*, $concurrent: \mathcal{AL} \times \mathcal{G} \rightarrow 2^{\mathcal{AT}}$ where,

$$concurrent(AL, g) = \{at \mid (at, g) \in AL\} \tag{2}$$

An agent completes each of its allocated tasks to a quality. We make the assumption that an agent which has been allocated multiple tasks must share its resources amongst those tasks until they are completed. This will reduce the quality of completion of those tasks as the number of concurrent allocations, $|concurrent(AL, g)|$, increases (See Example 3.7).

*Definition 3.6 (Atomic task quality).* The *atomic task quality* for an agent completing a task depends on the tasks' type, and the agents' concurrent allocations, $\omega_g: \mathcal{AP} \times \mathbb{N}_0 \rightarrow \mathbb{R}$ Therefore, the *allocation quality* of a set of tasks $AT$ given an allocation $AL$ is given by the mapping $ql: \mathcal{AT} \times \mathcal{AL} \rightarrow \mathbb{R}$,

$$ql(AT, AL) = \sum_{\forall at \in AT, (at, g) \in AL} \omega_g(type_a(at), |concurrent(AL, g)|) \tag{3}$$

*Example 3.7 (The effect of concurrency on task quality).* When completed independently, an agent in an ocean-deployed WSN can complete an oxygen sensing task $at_{oxy}$ in a time $t_{oxy}$, and a salinity sensing task $at_{sal}$ in a time $t_{sal}$, if it dedicates 100% of its resources, battery power and CPU cycles to the sampling, processing, and transmission of the results. The quality of these tasks in this system is based on how quickly the results can be returned. As the agents resources are finite, when the tasks are allocated concurrently the agent must split the required resources between the two executions, increasing the time to complete them. So, given an allocation $AL$ representing all of the current measurement tasks being carried out by nodes in the WSN,

$$ql(\{at_{oxy}\}, AL') + ql(\{at_{sal}\}, AL) > ql(\{at_{oxy}, at_{sal}\}, AL)$$

We make the assumption that the quality of composite task completion is simply the sum of the qualities returned from its allocations of atomic tasks. This is reasonable if there is minimal or zero effect on quality due to variations in a parent agents' aggregation of atomic tasks for different composite tasks. Also that the source of a parent agents' composite tasks does not effect the quality of its completion.

Over a period of time the system will progress through a number of system states as it allocates and completes tasks. If we define the mapping $tasks: \mathcal{AL} \rightarrow 2^{\mathcal{AT}}$ as the set of all tasks included in an allocation, such that, $tasks(AL) = \{at \mid (at, g) \in AL\}$, then we can define the utility of the system.

*Definition 3.8 (System utility).* The *utility* of a system is the sum of allocation qualities of each allocation in a set of system states, $u \colon \mathcal{S} \to \mathbb{R}$ so that,

$$u(S) = \sum_{(G, AL) \in S} ql(tasks(AL), AL) \tag{4}$$

*3.3.2 Optimality of allocations.* The range of allocations that an agent can achieve is bounded by its neighbourhood. The allocation of tasks $AT^{'}$ by agent $g$, to a neighbourhood of agents $G^{'}$ may be;

- non-optimal, there are other allocations of $AT^{'}$ to $G^{'}$ that will result in a higher allocation quality.
- locally-optimal, the allocation achieves the highest quality possible given the atomic tasks $AT^{'}$ and agents $G^{'}$.
- system-optimal, there is no other neighbourhood in the system that $g$ could have that would produce a higher quality given the tasks $AT^{'}$.
- non-allocable, the agents in the existing neighbourhood do not have the necessary capabilities to complete one or more of the tasks in $AT^{'}$.

*Definition 3.9 (Locally-optimal allocation).* The *locally-optimal allocation* of tasks $AT$ to agents $G$ given a fixed allocation of other tasks, $AL$, is the allocation that gives the highest quality $oq \colon 2^{\mathcal{AT}} \times 2^{\mathcal{G}} \times \mathcal{AL} \to \mathbb{R}$,

$$ol(AT, G, AL) = \underset{AL' \in permutations(AT, G)}{argmax} ql(tasks(AL'), AL \cup AL') \tag{5}$$

The quality of this locally-optimal allocation is then $oq \colon 2^{\mathcal{AT}} \times 2^{\mathcal{G}} \times \mathcal{AL} \to \mathbb{R}$,

$$oq(AT, G, AL) = ql(AL') \text{ where } AL' = ol(AT, G, AL) \tag{6}$$

Given the agents in the system, we can define all possible neighbourhoods for a given agent as $allhoods \colon \mathcal{G} \times 2^{\mathcal{G}} \to 2^{2^{\mathcal{G}}}$, where,

$$allhoods(g, G) = \{GS \in 2^{G} \mid |GS| < \delta_N\} \text{ given } g = (..., \delta_N, ...)$$

Of all these possible neighbourhoods, there will be one or more that will give the locally-optimal allocation with the highest quality within the system for an agents' allocation of a set of tasks.

*Definition 3.10 (Optimal neighbourhood).* In a system containing agents $G$ with allocation $AL$, the *optimal neighbourhood* of a set of atomic tasks $AT$, allocated to an agent $g$, will be that which gives the best quality, $on \colon 2^{\mathcal{AT}} \times \mathcal{G} \times 2^{\mathcal{G}} \times \mathcal{AL} \to 2^{\mathcal{G}}$, where,

$$on(AT, g, G, AL) = \underset{G' \in allhoods(g, G)}{argmax} oq(AT, G', AL) \tag{7}$$

The *system-optimal allocation* is therefore the locally-optimal allocation to this neighbourhood,

$$os(AT, g, G, AL) = ol(AT, on(AT, g, G, AL), AL) \tag{8}$$

Where the quality of this allocation will be,

$$osq(AT, g, G, AL) = ql(AL') \text{ where } AL' = os(AT, g, G, AL) \tag{9}$$

If we had full knowledge of the systems' tasks $AT$ and agents $G$, we could then calculate the *joint-optimal allocation*, $joq \colon 2^{\mathcal{AT}} \times 2^{\mathcal{G}} \to \mathcal{AL}$, the global allocation with the highest quality.

$$joq(AT, G) = oq(AT, G, \{\})$$

Our aim is to optimise towards this value using distributed algorithms, where only the agents' local-knowledge is used.

*Example 3.11 (Optimal allocations in multi-agent systems).* In Figure 1 we illustrate locally optimal, system-optimal, and joint-optimal allocations in a simple multi-agent system. A parent agent, $pg_1$, allocates two tasks, $\{at_1, at_2\}$, with type $\{ap_1, ap_2\}$ in a system with 3 child agents $cg_1$, $cg_2$ and $cg_3$. Each agent can complete tasks of these types to different task qualities $\omega_g(ap, |concurrent(AL, g)|)$.

If the neighbourhood of $pg_1$ is $N_1$, then, due to the effect of concurrency on each child agents' task completion qualities, the locally-optimal allocation for $pg_1$ in that neighbourhood is $\{(at_1, cg_1), (at_2, cg_2)\}$, with value 6. If $pg_1$ changed to neighbourhood $N_2$, then the quality of its task allocations to child agents $cg_2$ and $cg_3$ is the best possible in the system (if it were the only agent allocating tasks). So $N_2$ is its optimal neighbourhood for $pg_1$ given these tasks to allocate, and its system-optimal allocation would be $\{(at_1, cg_3), (at_2, cg_2)\}$, with quality 9. However, in a multi-agent system, parent agents may independently choose to allocate tasks to shared child agents. If $pg_2$ were also to allocate a task $at_1'$ of type $ap_1$ to $cg_3$, concurrency would cause the quality returned to $pg_1$ for $at_1$ to drop to 1, so the actual joint-optimal allocation[2] in this situation is $\{(at_1, cg_1), (at_2, cg_2), (at_1', cg_3)\}$, giving a quality of 11.

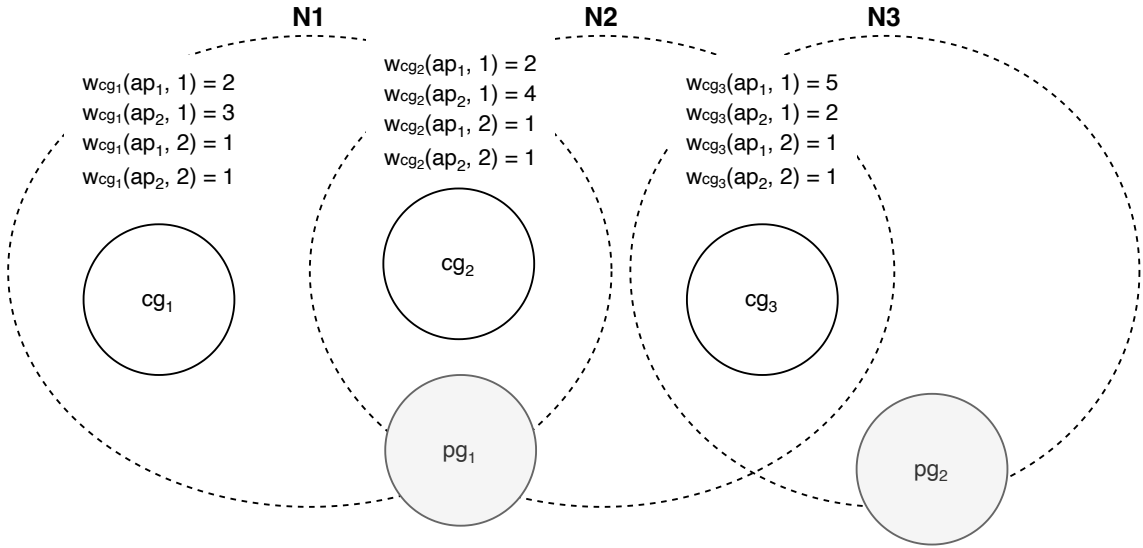

**Fig. 1.** Types of optimal task allocations in a multi-agent system.

---

[2]Note that multiple degenerate joint-optimal allocations may be possible in a system.

## 3.4 Finding the optimal task allocation

Given a set of agents and a set of composite tasks, how can we then find the joint-optimal allocation when agents' capabilities and task qualities are dynamic and unknown, therefore maximising the system utility? We separate this into two main sub-problems,

(1) Given a fixed local neighbourhood how can an agent find the locally-optimal allocation for an incoming distribution of composite tasks?
(2) How does an agent find the optimal neighbourhood within the set of all possible neighbourhoods it can achieve, containing the system-optimal allocation?

## 4 ALGORITHMS FOR OPTIMAL TASK ALLOCATION

We now give a high-level introduction to our algorithms for solving the task-allocation problem. The concepts and notation will be covered in more depth in Section 5.

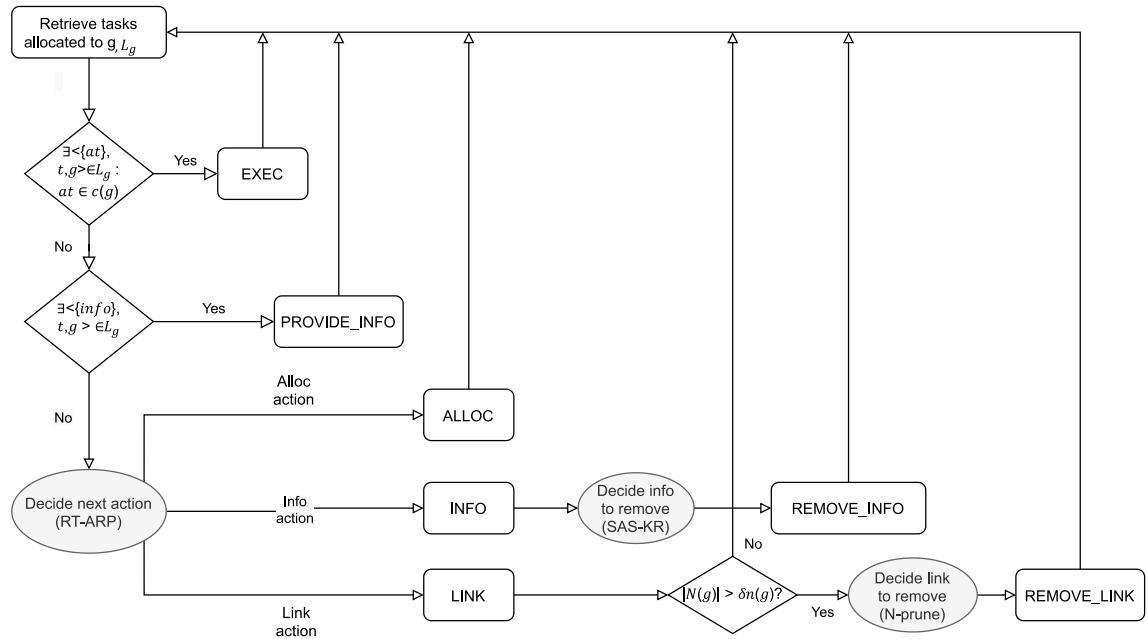

**Fig. 2.** Flowchart of the `ATA-RIA` algorithm. On receiving a composite task, an agent can carry out *EXEC* or *PROVIDE_INFO* actions immediately, or will choose amongst *ALLOC*, *INFO* and *LINK* using the `RT-ARP` algorithm. Taking an *INFO* or *LINK* action will lead to knowledge removal through the `SAS-KR` algorithm or neighbourhood pruning through the `N-Prune` algorithm respectively.

To tackle the problems as defined in Section 3.4, as well as handle the resource constraints of agents, we utilise two high-level approaches:

(1) agents should adapt their selection of actions based on their current performance, encouraging them to find better neighbourhoods within the system. Specifically, this adaptation is based on how an agent judges its own performance, given its neighbourhood and knowledge, and how good it believes this performance to be given the other possible neighbourhoods it has experienced.

(2) agents have constrained resources, and so must maintain a restricted neighbourhood and knowledge size. By retaining, and removing, neighbourhood agents and knowledge intelligently, they will be able to find neighbourhoods with better performance more efficiently than if they carried out the task randomly.

To achieve (1), we introduce the `RT-ARP` algorithm, for (2), we se the `N-Prune` and `SAS-KR` algorithms. To orchestrate the process and maintain the information needed for it to function, we use the `ATA-RIA` algorithm.

- the *agent task allocation with risk-impact awareness (*`ATA-RIA`*)* algorithm learns to take actions to optimise the task-allocation problem described. Its main purpose is to integrate the following three algorithms, as well as updating Q-values and sample data. It also selects actions based on measured progress towards composite task completion. (See Figure 2).

- the *reward trends for action-risks probabilities (*`RT-ARP`*)* algorithm increases the probability of an agent taking neighbourhood-altering actions and increasing exploration when the possible allocation quality achievable in its current neighbourhood is relatively poor compared to previous neighbourhoods.

- the *state-action space knowledge-retention (*`SAS-KR`*)* algorithm implements a knowledge retention scheme under dynamic neighbourhood changes. It removes parts of an agents knowledge less relevant to the optimisation problem so the agent can stay within resource bounds.

- the *neighbourhood update (*`N-Prune`*)* algorithm maintains an agents' neighbourhood within resource constraints by removing information on child agents based on their recent relative contribution to task completion quality.

### 4.1 The agent task allocation with risk-impact awareness (`ATA-RIA`) algorithm

The *agent task allocation with risk-impact awareness (*`ATA-RIA`*)* algorithm integrates the `RT-ARP`, `SAS-KR`, and `N-Prune` algorithms to provide a framework for optimising task-allocation in a multi-agent system (See Algorithm 1). It chooses between actions an agent can take. It then updates the Q-values of each action selected, based on the reward values returned using the reinforcement learning update algorithm described later in Section 5.2.3. We detail the steps when an agent is allocated a composite task below.

(1) execute an atomic task if the agent has the capability to do it [lines 2-6].

(2) otherwise choose an action based on `RT-ARP` [line 8].

(3) carry out the action and update the set of outputs, qualities, neighbours, and knowledge [lines 9-20].

(4) prune the knowledge base using `SAS-KR` to keep within the agents' resource bounds [line 17].

(5) prune the neighbourhood using `N-Prune` to keep within the agents' resource bounds [line 20].

(6) update the agents' Q-table mappings for the action taken and reward received using the *rlupdate* algorithm [line 22].

(7) update the action samples [line 24].

(8) repeat until all of the atomic tasks in the composite task are completed.

## 4.2 The reward trends for action-risks probabilities (`RT-ARP`) algorithm

The *reward trends for action-risks probabilities (*`RT-ARP`*)* algorithm judges the performance of an agents' current neighbourhood relative to previous ones using a `TSQM` (See Algorithm 2). It then takes the current Q-values for an agent and transforms them through the impact transformation function. The effect is to increase the probability of an agent taking neighbourhood-altering actions, and increasing the exploration factor, when the current neighbourhood is estimated to have a lower possible locally-optimal allocation than historical neighbourhoods. The steps are:

(1) select the agents' available actions and Q-values associated with the current state and Q-table. [line 1]

(2) generate an impact transformation function from the current `TSQM` and use it to transform the set of action to Q-value tuples,into action to likelihood tuples. [line 2].

(3) sum-normalise the resulting tuples to bound the values' sum to 1 and generate probabilities. [line 3].

(4) transform the exploration factor of the agent using the impact transformation function and use this for e-greedy action selection. This means more exploration when recent neighbourhoods have lower quality optimal allocations achievable [lines 4-5].

(5) either take the maximum-probability action or use random Boltzmann selection based on the transformed exploration factor. [lines 7-9].

## 4.3 The state-action space knowledge-retention (`SAS-KR`) algorithm

The *state-action space knowledge-retention (*`SAS-KR`*)* algorithm removes learned Q-values and knowledge based on the action information quality to stay within the bounds of an agents resource constraints (See Algorithm 3).

(1) find all an agents unavailable actions, those that involve agents that are in its knowledge base but not its neighbourhood [line 1].

(2) calculate the action information quality based on the staleness and amount of times actions have been taken [line 2].

(3) remove all knowledge of actions that have a value below a threshold value [line 4].

(4) remove all knowledge of another agent if there are no actions that target that agent [lines 5-6].

(5) check if the size of the knowledge base exceeds the constraint [line 9].

(6) remove a random agent from the knowledge base [line 10].

## 4.4 The neighbourhood update (`N-Prune`) algorithm

The *neighbourhood update (*`N-Prune`*)* algorithm ensures that an agents' neighbourhood is maintained at a size that bounds it within resource constraints (See Algorithm 4). Each child agents' contribution to task quality values are summed. Decay is used to reduce the relevance of older values. The information on the agents with the lowest contribution is then removed.

(1) compare the neighbourhood size with the resource limits [line 1].

(2) if the neighbourhood is too big and we have accumulated some quality values then select the agent that has produced the poorest quality value returns and remove it from the neighbourhood [lines 2-3].

(3) if the neighbourhood is too big and there are no quality values available then remove a random agent [line 5].

**Algorithm 1: The agent task allocation with risk-impact awareness (ATA-RIA) algorithm**

**Input:** $g$, The agent allocated the composite task
**Input:** $ct$, The composite task allocated to the agent
**Input:** $AT^\ominus$, The composite tasks currently unallocated atomic tasks
**Input:** $Q_g$, the Q-table mappings for agent $g$
**Input:** $W$, The potential change on neighbourhoods on taking an action.
**Input:** $\Lambda_g$, the TSQM matrix of summarised reward trends for agent $g$
**Input:** $\alpha$, a value $\mathbb{R}_{>0}[0, 1]$, weighting the rate of Q-value update
**Input:** $\lambda$, a value $\mathbb{R}_{>0}[0, 1]$, weighting importance of future rewards
**Input:** $\hat{\mu}_{\min}$, The information retention threshold.
**Input:** $SP$, the set of action samples

**Result:** $N(g)$, updates to the neighbourhood of agent $g$
**Result:** $K(g)$, updates to the knowledge base of agent $g$
**Result:** $Q'_g$, updates to the Q-table of agent $g$
**Result:** $SP'$, updates to the set of action samples

```
1  for at ∈ ct do
      // Execute atomic task if agent has capabilities
2     if category(a) ∈ c(g) then
3        EXEC(g, at)
4        if at is successfully completed then
5           ct ← ct − {at}
6        end
7     else
         // Select an action given unallocated tasks
8        a ← RT-ARP(type_a(AT^⊖), W, Λ, ε_base)
9        if category(a) = ALLOC(g, at, n) then
10          ALLOC(g, at, n)
11          if at is successfully completed then
12             ct ← ct − {at}
13          end
14       else if category(a) = INFO(g, t, n) then
            // Get new agent k from action
15          k ← INFO(g, t, n)
16          K(g) ← K(g) ∪ {k}
            // Prune knowledge base
17          SAS-KR(AT^⊖, N(g), K(g), s(SP, actions(A, g)), μ̂_min)
18       else if category(a) = LINK(g, k) then
            // Add new agent to neighbourhood
19          LINK(g, k)
            // Prune neighbourhood based on resources
20          N-Prune(N(g), SP)
21       end
         // Update Q-value mappings using reward generated by action
22       Q'_g ← rlupdate(Q_g, type_a(AT^⊖), a, r)
         // Use the quality value to update the TSQM
23       updatetsqm(Λ_g, ω)
         // Update action samples
24       SP' ← SP ∪ {(a, t, ω)}
25    end
26 return (N(g), K(g), Q'_g, SP')
```

## 5 OPTIMISATION USING REINFORCEMENT LEARNING-BASED TASK ALLOCATION

Next we detail the concepts and definitions that are used within our algorithms. We see how the probability of agents taking different types of actions can be changed based on previous experiences. The awareness of their possible impact is also an important aspect in predicting whether certain actions will increase or decrease the likelihood of agents achieving optimal allocation solutions. We first look at how to quantify the value of information in a neighbourhood,

---

**Algorithm 2: The reward trends for action-risks probabilities (RT-ARP) algorithm**

**Input:** $AT^{\ominus}$, the set of unallocated atomic tasks of agent $g$
**Input:** $Q_g$, the Q-table mappings for agent $g$
**Input:** $W$, the action-risk values for the available actions
**Input:** $\Lambda$, the TSQM used to generate the transformation function
**Input:** $\epsilon_{\text{base}}$, the base exploration factor for the learning algorithm

**Result:** $a$, the action for the agent to carry out

// Select the available action to Q-value tuples associated with the agents' current state
1   $AQ \leftarrow available(Q_g, type_a(AT^{\ominus}))$
   // Scale the action to Q-value tuples element-wise using impact-transformation
2   $AQ \leftarrow (AQ \circ it(W))$
3   $AQ \leftarrow \text{sumnorm}_p(AQ)$
   // Calculate the impact exploration factor
4   $\epsilon_{\text{ief}} \leftarrow it(0.5)$
   // Scale the base exploration value
5   $\epsilon \leftarrow \epsilon_{\text{base}} \times \epsilon_{\text{ief}}$
   // Select best action or explore with boltzmann selection
6   **if** $\text{rand}(\mathbb{R}[0,1]) < \epsilon$ **then**
7      $a \leftarrow \max_b(AQ)$
8   **else**
9      $a \leftarrow \text{boltzmann}_b(AQ)$
10   **end**
11   **return** $a$

**Table 1.** Summary of standard functions

| Function | Definition | Summary |
|---|---|---|
| $\text{sumnorm}_p(Q)$ | $Q \leftarrow \left\{ \left(a_i, \frac{p_i}{\sum_{j=1}^{N} p_j}\right) \right\}_{\forall (a_i, p_i) \in Q}$ | *Sum normalisation*, scales $p$ values in a set $Q = \{(a_i, p_i)\}_{i=1}^{N}$ uniformly into the range $\mathbb{R}[0,1]$, where the resulting $p$ values sum to 1. |
| $\text{softmax}_p(Q)$ | $Q \leftarrow \left\{ \left(a_i, \frac{e^{p_i}}{\sum_{j=1}^{N} e^{p_j}}\right) \right\}_{\forall (a_i, p_i) \in Q}$ | *Softmax normalisation*, scales $p$ values in a set $Q = \{(a_i, p_i)\}_{i=1}^{N}$, into the range $\mathbb{R}[0,1]$. |
| $\text{rand}(Q)$ | $a \xleftarrow[P(X)]{} Q, \ P(X) = \left\{\frac{1}{|X|}\right\}$ | *Uniform selection*, Selects a value $a$ in set $Q = \{(a_i)\}_{i=1}^{N}$ using the uniform distribution. |
| $\max_b(Q)$ | $a \leftarrow \text{argmax}_b Q$ | *Maximum selection*, returns a value $a$ in set $Q = \{(a_i, b_i)\}_{i=1}^{N}$ with the maximum value of $b$. Randomly selects between degenerate values. |
| $\text{boltzmann}_b(Q)$ | $a \xleftarrow[P(X)]{} Q, \ P(X) = \left\{\frac{e^{(b_i/\tau)}}{\sum_{j=1}^{N} e^{(b_j/\tau)}}\right\}_{\forall (a_i, b_i) \in Q}$ | *Boltzmann selection*, returns a value $a$ in set $Q = \{(a_i, b_i)\}_{i=1}^{N}$, chosen using the Boltzmann distribution of $b$ values with absolute temperature value $\tau$. |

then how to adapt Q-learning for a non-stationary environment. This helps in predicting how different actions will impact an agents' knowledge and neighbourhood, guiding them towards optimal parts of state-space. We use these predictions to judge the performance of an agents' current policy as compared to its previous ones, then determine whether it should take actions that it currently predicts to be non-optimal, or actions that will substantially change its neighbourhood or knowledge base, in order to explore the action-space and achieve better performance. In combination, these approaches allow agents to adapt as the optimal policy of a non-stationary system changes.

## 5.1 The value of information and neighbourhoods

*5.1.1 Action samples.* In order to use the agents' historical performance to alter its future behaviour we need to store information on past actions and their outcomes. We do this through *action sample* tuples $sp = \langle a, t, \omega \rangle$, where $a$ is an action taken at time $t$ that gave quality $\omega$. We define the *action sample selection* function to allow us to specify subsets

---

**Algorithm 3: The state-action space knowledge-retention (SAS-KR) algorithm**

---

**Input:** $AT^\ominus$, the set of unallocated atomic tasks of agent $g$
**Input:** $N(g)$, the neighbourhood of agent $g$
**Input:** $K(g)$, the knowledge base of agent $g$
**Input:** $SP$, the set of action samples
**Input:** $Q_g$, the Q-table mappings for agent $g$
**Input:** $\tilde{\mu}_{\min}$, The information retention threshold.

**Result:** $K(g)$, updates to the knowledge of agent $g$
**Result:** $SP'$, updates to the action samples
**Result:** $Q'_g$, updates to the Q-mappings of agent $g$

```
   // Select all unavailable actions
1  for (a, q) ∈ unavailable(Q_g, type_a(AT^⊖)) do
      // Test the action meets the information retention threshold
2     if mv(SP, a, t) < μ̃_min then
         // Remove all samples of action a
3        SP' ← SP − {(a, t, ω) | (a, t, ω) ∈ s(SP, actions(A, g))}
         // Remove actions and learned Q-values
4        Q'_g ← Q_g − {a}
         // Remove agents in g's knowledge that are not targets of any action in Q
5        X = {x | ∀(a, p) ∈ Q, x ∈ K(g), a ∉ targets(A, g, K(g))}
6        K(g) ← K(g) − X
7     end
8  end
   // Check if knowledge size exceeds resource limit
9  while |K(g)| > δ_k(g) do
      // Remove a random agent in the knowledge base but not neighbourhood
10    K(g) ← K(g) − rand(K(g) − N(g))
11 end
12 return (K(g), SP', Q'_g)
```

---

**Algorithm 4: The neighbourhood update (N-Prune) algorithm**

---

**Input:** $N(g)$, the neighbourhood of the agent.
**Input:** $SP$, The set of action samples

**Result:** $N(g)$, the updated neighbourhood of the agent.

```
   // Check if neighbourhood size exceeds resource limit
1  while |N(g)| > δ_n(g) do
2     if |s(SP, actions(A, g))| > 0 then
         // Find the neighbour agent that has returned the lowest total quality
3        n ← mvn(SP, g)
4     else
         // Choose a random neighbour agent
5        n ← rand(N(g))
6     end
      // Remove the neighbour agent
7     N(g) ← N(g) − {n}
8  end
9  return N(g)
```

---

of action samples for which the action performed was an element of a given set of actions, $A$.

$$s(SP, actions(A, g)) = \{(a, t, \omega) \mid \forall (a, t, \omega) \in SP, \exists a \in A\} \tag{10}$$

For convenience, we also define the *latest sample time* in a set of samples, $latest : \mathcal{SP} \to \mathbb{R}$, as a function that takes a set of actions samples returns the time of the most recent sample.

*5.1.2 Information value.* We first make an assumption that the more recently, and frequently, an agent takes an action, the better accuracy it has in predicting of that actions' contribution to longer-term rewards. This allows us to define the *action information quality*, a proxy for the value of information collected about an action $a$ at time $t$, given the set of action samples $SP$.

$$mv(SP, a, t) = \frac{|s(SP, actions(A, g))|}{t - latest(s(SP, \{a\}))} \tag{11}$$

The *uncertain information threshold* $\hat{\mu}_{\min}$ is then chosen as the minimum value below which an agents' information about the expected rewards of taking an action is no longer considered useful for prediction.

*5.1.3 The value of neighbourhood agents.* We define *neighbour information value* as the sum of the quality values of all action samples $SP$ of an agent $g$ that refer to actions that involve agents in a set $G$.

$$nval(SP, g, G) = \sum_{(a,t,\omega) \in SP, \; a \in targets(A,g,G)} \omega \tag{12}$$

*Definition 5.1 (Minimum value neighbour).* The *minimum value neighbour* of an agent $g$ is the child agent that generates the least neighbour information value.

$$mvn(SP, g) = \underset{x \in N(g)}{argmin} \; nval(SP, g, \{x\}) \tag{13}$$

## 5.2 Adapting Q-learning techniques for non-stationary environments

For all possible actions an agent can take there is a likelihood that taking that action in the current state will increase future rewards. When an action is taken, the accuracy of these values can be improved based on the actual rewards returned. This is how an agent can improve its *policy*, the mapping of its view of the state of the system, and the actions it can take within those states, to the likelihood those actions are optimal. Q-learning methods have been successfully applied as a model-free method of learning policies [16], however, in real-world multi-agent systems, we commonly find state-spaces that are only partially-observable to agents, and are non-stationary, where Q-learning is more complex to apply [17, 18].

*5.2.1 Strategies for policy change in non-stationary environments.* In stationary environments the agents' policy targets a single Markov Decision Process (MDP) that does not change [78, 86]. Over time, actions an agent believes to be non-optimal become increasingly unlikely to be updated. With multiple agents interacting however, the environment is now non-stationary, and becomes an infinite random sequence of MDPs [8]. Without sampling actions previously judged to be non-optimal, an agents' policy will not adapt. Unless it can detect changes in the sequence of MDPs, and explore new actions, it becomes stuck applying a previous policy to increasingly different MDPs [69].

To adapt reinforcement learning algorithms for these systems, we can replay action updates by updating the learning algorithms with the reward for a previously taken action once more , with the repetition frequency being inversely proportional to their likelihood of being chosen [2], or reset values periodically to restart the learning process [58]. However, these approaches do not take account of changes in the rate of drift of the optimal policy during the systems' lifetime, or if the change is transient (e.g. temporary weather conditions). Our algorithm tackles this problem in the following ways;

(1) An agent will optimise for its partially observable state using reinforcement learning by default (see Section 5.2.3).
(2) If current rewards are good compared to historical values, the `ATA-RIA` algorithm assumes a relatively stationary state, and continues reinforcement learning updates to optimise the agents' policy.
(3) If rewards are historically poor, the `RT-ARP` algorithm will increase the likelihood of choosing non-optimal actions, updating them more frequently. This likelihood, and how it varies between action-categories depending on the scale of their impact on an agents' policy, changes with the scale of the difference with those historical rewards (see Sections 5.3- 5.5).
(4) As the policy changes, previous knowledge becomes less valued, and is forgotten, by the `SAS-KR` algorithm, gradually resetting learning.

*5.2.2 Q-learning state, actions, and rewards.* For each agent, we define a Q-table mapping, $Q_g$, where the agent will learn the maximum expected future rewards for each action, the Q-value, in each state, allowing the agent to choose the best action for this state in the future. A state in this table, $s_t$, will be the set of atomic task types that the agent is still to allocate. In each state, there is a choice of actions, $a_t$. The agent could carry out an *ALLOC* action for every unallocated task type, where each allocation could be to any of the agents in its neighbourhood. The agent could carry out an *INFO* action type, or it could carry out a *LINK* action type. On carrying out an action, the agent obtains a reward $r_t$. For *ALLOC* actions this is simply the quality of task completion by the agent the task is allocated to. For *INFO* and *LINK* actions we set a fixed, slightly negative reward value.

*5.2.3 Updating a Q-table.* Q-table values are updated using standard Q-learning [82], where the learning rate $0 \leq \alpha \geq 1$ controls the step-size of the update, and a discount factor $0 \leq \gamma \geq 1$ alters the value of rewards depending on how recent they are.

$$Q^{new}(s_t, a_t) \leftarrow \underbrace{Q(s_t, a_t)}_{\text{old value}} + \underbrace{\alpha}_{\text{learning rate}} \cdot \underbrace{\left( \underbrace{r_t}_{\text{reward}} + \underbrace{\gamma}_{\text{discount factor}} \cdot \overbrace{\underbrace{\max_a Q(s_{t+1}, a)}_{\text{estimate of optimal future value}}}^{\text{temporal difference}} - \underbrace{Q(s_t, a_t)}_{\text{old value}} \right)}_{\text{new value (temporal difference target)}} \quad (14)$$

This update process ignores the possibility that the optimal action policy can change, however as discussed, this is mitigated by the other steps described in Section 5.2.1 and detailed in the following sections.

To help simplify our algorithm definitions, we define two functions. The *rlupdate* : $Q_g \times \mathcal{AP} \times A \times \mathbb{R} \rightarrow Q_g$ function takes an agents' current Q-table, a set of unallocated atomic tasks, an action, and a reward, and returns the updated Q-table. Whereas the *rlselect* : $Q_g \times \mathcal{AP} \rightarrow \mathcal{A} \times \mathbb{R}$ function, returns a set of tuples of actions and their Q-values given a Q-table and a set of unallocated atomic task types to select the state.

*5.2.4 The availability of actions.* In a given state, not all of the actions an agent knows of are available for it to take. For example an agent $g$ cannot perform an allocation action ALLOC($g, at, n$) if $n \in K(g)$ but $n \notin N(g)$. We refer to these as *unavailable actions, unavailable* : $Q_g \times \mathcal{AP} \rightarrow \mathcal{A}$, actions that involve other agents in an agents' knowledge base, but are not currently in its neighbourhood. An agents' set of *available actions*, *available* : $Q_g \times \mathcal{AP} \to \mathcal{A}$, are the actions it can take given its unallocated atomic tasks, neighbourhood, and knowledge.

### 5.3 Predicting the effect of actions

Different action-categories change an agents' neighbourhood, $N$, or knowledge base, $K$ to differing degrees. Predicting how, and to what extent, each action will impact an agents' current policy is key to adapting Q-learning to handle the non-stationary environment.

To enable agents to make these predictions we,

(1) Define the impact of the different categories of actions on both an agents' neighbourhood and knowledge.
(2) Estimate the probability that actions generating impact will actually occur.
(3) Combine these factors to define action impact.
(4) Detail algorithms based on historical quality values to predict which action impacts will have a positive effect on task completion quality.

*5.3.1 Neighbourhood and knowledge impacts.* There is an impact on possible allocation quality if an agent takes actions that change its neighbourhood. This *neighbourhood impact* on an agent allocating atomic tasks $AT$ from changing its neighbourhood from $N'$ to $N''$ within a system with allocation $AL$ is the difference between the locally-optimal allocation qualities of the respective neighbourhoods,

$$ni(AT, N', N'', AL) = oq(AT, N'', AL) - oq(AT, N', AL) \tag{15}$$

*Definition 5.2 (Maximum neighbourhood impact).* The *maximum neighbourhood impact* is the maximum possible neighbourhood impact given a set of atomic tasks $AT$ and all combinations of neighbourhoods that can be formed from a set of agents $G$.

$$mni(AT, G, AL) = \underset{\forall (X,Y) \subseteq (2^G \times 2^G)}{argmax} ni(AT, X, Y, AL) \tag{16}$$

*Definition 5.3 (Knowledge impact).* The *knowledge impact* of an agent changing its knowledge from set of knowledge $K'$ to $K''$ is the difference between the maximal neighbourhood impacts.

$$ki(AT, K', K'', AL) = mni(AT, K'', AL) - mni(AT, K', AL) \tag{17}$$

*Example 5.4 (Impact).* An agent $g$ in a marine WSN system has a knowledge base from which it can form 3 distinct neighbourhoods, $N_1, N_2$ and $N_3$, and is currently allocating oxygen reading tasks, $ap_{oxy}$ to $N_1$. The locally-optimal allocation quality of $N_2$ is worse than that of $n_1$ (for example, due to low battery levels), and that of $N_3$ much better. In this case, if $g$ was to take an action to replace $N_1$ with $N_2$, then this would give $ni(\{at_{oxy}\}, N_1, N_2, AL) < 0$, a negative impact. In contrast, taking an action that replaces $N_1$ with $N_3$ would give $ni(\{at_{oxy}\}, N_1, N_3, AL) > 0$, which is then the maximum neighbourhood impact, given the knowledge base $N_1 \cup N_2 \cup N_3$.

*5.3.2 The probability of impact effects.* Since neighbourhoods and knowledge are dynamic, agents are continually added and removed. Therefore there is a probability that agents in a neighbourhood never contribute to the quality of a composite task before they are removed or the task is completed. In other words, when an agent moves from a neighbourhood $N'$ to $N''$, it will lose access to actions involving agents in the set $N' - N''$, and gain those in the set $N'' - N'$. If actions due to agents in those sets are never taken, there is no overall impact to allocation qualities on changing the neighbourhood or knowledge base. We define the probability an agent takes actions involving its neighbourhood agents in the sets $N' - N''$ or $N'' - N'$ as the *neighbourhood impact probability*, $P(N', N'')$. Similarly we define the *knowledge impact probability* as $P(K', K'')$.

*5.3.3 Estimating the impact of taking an action.* The *action impact* is the expected value of the change in possible allocation quality if an action $a$ is taken. On taking the action the neighbourhood is changed from $N' \rightarrow N''$ and the knowledge base from $K' \rightarrow K''$.

$$ai(AT, N', N'', K', K'', AL) = P(N', N'')ni(AT, N', N'', AL) + P(K', K'')ki(AT, K', K'', AL) \tag{18}$$

 The probability an agent will take different actions, and how those types of actions will impact the possible qualities of the atomic task completions in a changed neighbourhood or knowledge base, are generally unknown, and difficult to predict, in complex, dynamic systems. Using estimations allows us to focus on the core algorithm behaviours without adding complexity. In order to estimate action-impact values we make some large simplifications. As these values are combined with a learned value the `RT-ARP` algorithm, the important quality of these approximations is that the different types of action are separated, and that the relative impact ordering of them is the same as would be with an accurate calculation. As long as these properties are maintained, the algorithms should work as expected, the more accurate the estimations, the more quickly we should expect learning to proceed. However, implementing more granular impact prediction in future work could be expected to improve the algorithms' performance.

As a first approximation, we estimate action-impacts $\widehat{ai}$, based on whether each action-category changes the state of neighbourhoods or knowledge bases, and the probability of the impacts described given a systems' size. We detail how these estimations are calculated in Appendix C.

*Action-impact values $W$* are estimated values for maximum action impacts for each action-category, $category(a)$. We assume that both $|N'' - N'| \in \{0, 1\}$ and $|K'' - K'| \in \{0, 1\}$ for all actions. We also assume that $AL_S$ is large enough to remain approximately constant despite any allocation change or resource pressure resulting from the action.

$$W = \{(category(a), \widehat{ai}(AT, N', N'', K', K'', AL)) \mid \forall a \in A\} \tag{19}$$

## 5.4 Measuring relative allocation optimality

For an agent to know the optimal task quality it could achieve in its current neighbourhood we use a metric to measure how far its current quality values are from optimal.

*Definition 5.5 (Locally-optimal allocation metric).* The *locally-optimal allocation metric* is the difference between an agents' current allocation quality of atomic tasks $AT$ to agents in its neighbourhood, and the locally-optimal allocation quality.

$$d_{\texttt{loc}}(AT, g, AL) = oq(AT, n(g), AL) - ql(AT, AL) \tag{20}$$

*Definition 5.6 (System-optimal allocation metric).* The *system-optimal allocation metric* is the difference between an agents' current allocation quality and the system-optimal allocation quality given the set of agents in the system, $G$.

$$d_{\texttt{sys}}(AT, g, G, AL) = osq(AT, g, G, AL) - ql(AT, AL) \tag{21}$$

### 5.5 Predicting impact from historical performance

An agent needs to know the locally-optimal allocation quality for both the current and the future neighbourhoods to predict whether the impact of changing neighbourhoods from $N'$ to $N''$ would be positive. This is difficult since the agent is uncertain of $d_{loc}$ and so does not know the best values it can obtain in the current neighbourhood. However, it is likely to have less samples of the actions available in $N''$ so may have even more uncertainty in future values if it changed neighbourhoods. To find proxies for these values we make the following assumptions based around time-based trends in action-samples.

ASSUMPTION 1. *(Likelihood of neighbourhood change) The more actions an agent takes the greater the likelihood that it will have taken actions that change its neighbourhood.*

If there is always some exploration of the action-space, making this assumption reasonable. In our algorithms we utilise Boltzmann selection so this holds true. In some annealing-based learning algorithms, exploration of the action space will decrease over time and this may not hold true. However, in dynamic systems these non-adaptive, time-based approaches would not be applicable in any case.

ASSUMPTION 2. *(Variation of neighbourhoods) Samples in a large set of historical action-samples will come from many different neighbourhoods.*

Making this assumption allows an agent to compare its current performance with historical values and assume it represents a statistical comparison of its current neighbourhood to others in the system. Where an agent has access to all its possible neighbourhoods, our algorithms should find the system-optimal neighbourhood for that agent. If the agent can only access a small subset of neighbourhoods, it should find the best in that subset. As such, the algorithms should perform well in both scenarios.

ASSUMPTION 3. *(Time-dependent similarity of neighbourhoods) Action-samples separated by short spaces of time are likely to be from similar neighbourhoods. Those separated by large amounts of time are more likely to represent very different neighbourhoods*

Using this assumption, if an agent has had much better rewards in the past, it can use these to infer that its current neighbourhood might benefit from being substantially changed to improve performance. However, in systems where agents can only reach a very limited possible set of neighbourhoods (e.g. due to their static location and limited broadcast range), this may not be reasonable. In such systems our algorithms would push agents to take risky actions that substantially alter their neighbourhoods and knowledge, when exploitation of the current one may be the better choice.

By making these assumptions we can estimate the relative locally-optimal allocation and system-optimal allocation metric values. As recent action-samples with small time separations come from the same or similar neighbourhoods we compare their quality value statistics to estimate $d_{\text{loc}}$. As action-samples over the long-term come from many different neighbourhoods, we compare their values to estimate $d_{\text{sys}}$.

*5.5.1 Methods to estimate action-impacts.* To estimate which actions will have a positive impact we firstly use historical action-sample quality values to estimate action-impacts. Based on these values we increase or decrease the probabilities of taking different action-categories. Whether an impact is estimated to be positive or negative will alter the agents likelihood of taking actions that explore allocation within the current neighbourhood or change its neighbourhood or knowledge base. The process is as follows,

(1) define the *time-summarised quality matrix (TSQM)*, a method of summarising historical quality returns over multiple time scales. This uses a resampling technique where each row in the matrix is the result of downsampling the time-series data of the previous row [79]. The update period for each rows' recalculation, and the frequency of downsampling is dictated by the matrix dimensions, as detailed in Section 5.5.2.

(2) using this matrix we generate the *impact interpolation function*.

(3) we then define the *impact transformation function* using a ratio of the integrations over the impact interpolation function.

(4) finally we use the action-impact values for each action-category that that will be used to as the input for the impact transformation function.

*5.5.2 Time-summarised quality matrix (`TSQM`).* A `TSQM` $\Lambda$ has shape $(m \times n)$ with all values initially null. Time-ordered actions-sample quality values $\{\omega_t, \omega_{t-1} \ldots, \omega_{t-n}\}$ for all actions of a specific agent are added to the first row $\Lambda_{(0,j)}$ as they are sampled such that, $\Lambda_{(0,)} \leftarrow \{\omega_i\}_{i=0}^{n}$. Each subsequent row is the result of averaging and pooling values in the previous row. This approach allows each row to represent the quality trends across different time-scales. If $h$ is the number of quality values added to the matrix then we update the elements as follows,

$$\Lambda_{(i+1,k)} \leftarrow \frac{\sum \Lambda_{(i,)}}{|\Lambda_{(i,)}|}, \text{ if } h \bmod (k |\Lambda_{(i,)}|) = 0 \tag{22}$$

To summarise the process of updating the matrix,

- each new value of $\omega$ updates the first cell of the initial row of $\Lambda$, row 0. As this cell is updated, the other values in the row are moved along by one to accommodate it, with the last value being discarded.
- after $n$ new values of $\omega$ have been added to row 0, an average of row 0 is taken and added as the first cell of the row 1. All values on row 1 are moved along by one, and the last value discarded. The same process will happen after the addition of each batch of $n$, $\omega$ values to row 0.
- after $n$ new average values have been added to row 1, row 2 will have the average of row 1 added to its first cell, moving all the others along and discarding the last.
- the same process continues for all other rows. Where each will update its first cell with the average of the previous row after $n$ values have been added to that row.

We use the function `updatetsqm`$(\Lambda_g, \omega)$ as shorthand for the full update process for an agent $g$.

Note that the values of $(m \times n)$ will alter the behaviour of the `TSQM` in the following ways. Increasing the value of $m$ will increase the number of trends over different timescales that the agent will use. Whereas the larger the value of $n$, the greater the number of quality values the agent must receive before it updates each of these longer-term trends.

*5.5.3 Impact interpolation function.* The *impact interpolation function* $ii(x)$ is generated by taking a linear interpolation[3] over the rows of a `TSQM` (see Figure 3). A decay factor $\delta \in [0,1]$ is chosen to dampen the values of longer time-scales (exponentially by the row exponent $i$) to allow more recent trends to have a stronger impact. For a `TSQM` of shape $(m \times n)$ a value $x \in \mathbb{R}[0,1]$ will be transformed as below.

$$ii(x) = \text{interpol}\left[\left\{\left(\frac{i}{N}, \text{average}(\Lambda_{(i,)})\delta^i\right)\right\}_{i=0}^{N}\right](x), \text{ for layers 0 to } N \tag{23}$$

---

[3]We use a 1-d linear interpolation, implemented using the python `scipy.interpolate.interp1d` method. Where interpol$[\{(x_1, y_1), (x_2, y_2), ..., (x_n, y_n)\}](x) = y$ estimates the value $y$, from $x$, using an interpolated function generated from a known set of values, $\{(x_1, y_1), (x_2, y_2), ..., (x_n, y_n)\}$.

The effect of this is to dynamically generate a function from the TSQM matrix where a parameter $x$ will be mapped to $\omega$-value trends, with larger values of $x$ mapping to longer terms trends.

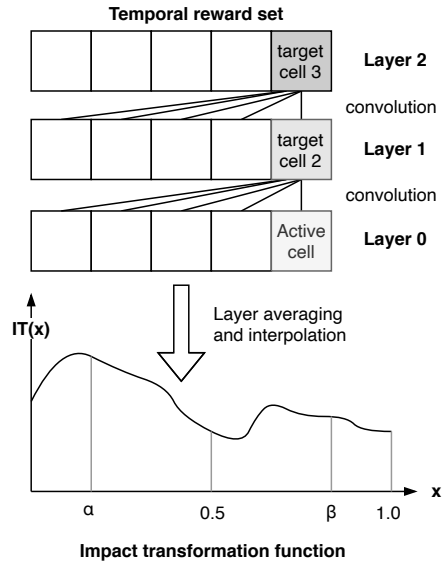

**Fig. 3.** Transforming the TSQM

*5.5.4 Impact transformation function.* The *impact transformation function* estimates the probability that taking an action from an action-category in the current neighbourhood will be positive by taking a ratio over the integrals of the interpolation representing the fraction of the historical quality values that occur up to the input value. For any $y \in \mathbb{R}[0, 1]$ this is given by,

$$it(x) = 1 - \frac{\int_{y=0}^{x} ii(y)\,dy}{\int_{y=0}^{1} ii(y)\,dy} \tag{24}$$

Given a value $x$, if $it(x)$ is close to 0, then the quality values generated by the system have been better than over the time period represented by the range $[x, 1]$, than over the shorter time period $[0, x]$. This shows the systems' near-term performance is worse than its longer-term performance. Conversely, if $it(x)$ is close to 1, then the short-term performance of the system is better than its previous longer-term trends.

We use this balance of the impact transformation function between shorter and longer timescales to adapt the exploration behaviour of our reinforcement learning model. We use the midway value of $x = 0.5$, as a *impact exploration factor* is defined as, $\epsilon_{ief} = it(0.5)$. Higher values mean the agent is attaining better performance now than in the past and should exploit rather than explore the system further. Lower values mean its exploration of the system should be increased.

Finally we can use the interpolation of action-impact values $w$ of action-categories of each action $a$ to estimate the probability that taking those type of action will have a positive impact,

$$P(ni(AT, N^{'}, N^{''}, AL) > 0 \mid a) \approx it(w) \tag{25}$$

Using this method, agents will prefer lower-risk actions when the system is performing well, $it(w) \rightarrow 0$. and higher-risk actions when the systems performing historically poorly, $it(w) \rightarrow 1$.

## 6 EVALUATION

### 6.1 Simulation

*6.1.1 Systems and algorithms.* We simulated four dynamic systems to evaluate the algorithms' performance. In all systems, to simulate realistic communication and environmental effects, each agent in the system had a 0.1% chance being unavailable for each episode. This value was chosen as a reasonable failure rate given the possible component failure modes of ocean-based WSN hardware, and the current and salinity effects that can disrupt communications [7, 19, 89, 92]. Each simulation was run 100 times. In the *stable system* we look at the performance of the `ATA-RIA` algorithm on the task allocation problem overall, when agents' neighbourhoods were randomly assigned on initialisation. The *exploration system* focuses on how the `RT-ARP` algorithm alters the probability of exploring system space to find the best neighbourhood for each agent. In this system we initialise parent agents' neighbourhoods to contain child agents with atomic task qualities that are more or less than the average in the system[4]. We then investigate how agents adapt these neighbourhoods to improve performance. The *volatile system* examines the adaptability of the algorithms when the system is highly dynamic, such as during transient environmental events, by randomly making 1% of each parent agents' neighbourhood agents unavailable per-episode during a defined period of disruption between episodes 25 and 75. This represents a $10x$ increase in failure rates as compared to the stable system, a value chosen to simulate the increased component and communication failures possible in a harsh environment such as an ocean-based WSN in rough seas. Finally, in the *large system* we look at scalability, the performance of the algorithms as we increase the number of agents in the system.

*6.1.2 Theoretical system optimal utility as a baseline comparison.* In a system with a single task to complete there will be an agent that can complete the task to the best or equal quality of all the available agents. With no resource sharing amongst tasks, the allocation of tasks within the system, *AL* can be ignored, and enumerating the possible solutions within the simulated systems is greatly simplified. We use this approach to give us the theoretical optimum utility in a system where atomic tasks are completed in isolation, which we then use as an easily computable comparison set of data for our simulation systems.

$$u^*(S) = \sum_{\forall (G_s, AL) \in S} ql(tasks(AL), \{\}) \tag{26}$$

In our simulations we have detailed knowedge of the state and the quality of task completion of all the agents in the system. With the removal of any resource competition these will not change with concurrency, and we can enumerate these at system initialisation to calculate $u^*(S)$ for each allocation during each episode. We can then use the utility loss w.r.t. the theoretical optimal as a baseline comparison, as used in Figures 4, 5 6, and 7.

*6.1.3 Comparison algorithms.* As comparisons, we implement two Q-learning based algorithms in the stable environment. The `<qlboltz>` algorithm uses the *RLUpdate* strategy shown in Section 5.2.3, with the addition of Boltzmann exploration (See Table 1). The temperature used to reduce exploration over the lifetime of the simulation [21] was the episode count. We also used the `<qlreset>` algorithm for comparison, based on work extending Q-learning to non-stationary systems [10, 13, 58] as described in Section 5.2.1. Extending the `<qlboltz>` algorithm, we add a simple

---

[4]Child agents' atomic task qualities were set at system start time from values in the range $(0, 1]$ drawn randomly from the *normal distribution* defined by values in $X \sim \mathcal{N}(\mu, \sigma^2), \mu = 0.5, \sigma = 0.2$

memory-resetting strategy that partially resets an agents' learned Q-values each episode by updating every value in the agents' Q-table with half the values' difference from the average for that state.

*6.1.4 Configuration of algorithms, systems, and data presentation.* Labels for the algorithms and configurations used in the simulations are described in Tables 2, 3, 4, and 5. System parameters are included in Appendix A, with general and individual system values shown in Tables 7, and Table 8 respectively. The TSQM uses $(m \times n)$ parameters of $(10 \times 10)^5$. The composite task frequency distribution introduced the same fixed set of tasks over a specified period, defining each *episode* of the system.

**Table 2.** Summary of labels for the stable system

| Label | Summary |
|---|---|
| <optimal> | This algorithm is used as a performance comparison as it provides the theoretical optimum system utility. Its parent agents are initialised with the most optimal neighbourhoods available in the system, and always allocate tasks to the highest quality child agents. |
| <qlboltz> | Q-learning algorithm with Boltzmann exploration. |
| <qlreset> | Q-learning algorithm with Boltzmann exploration and episodic reset of learned Q-Values. |
| <ataria> | The ATA-RIA algorithm. |

**Table 3.** Summary of labels for the exploration system

| Label | Summary |
|---|---|
| <rtrap$^0$> | ATA-RIA when the system is initialised with random neighbourhoods then explores with a constant $\epsilon$ factor, RT-ARP is disabled. This is used for a baseline comparison. |
| <rtrap$^+$> | ATA-RIA when the system is initialised with neighbourhoods containing 75% of the optimal neighbourhoods' agents and explores using RT-ARP. |
| <rtrap$^-$> | ATA-RIA when the system is initialised with neighbourhoods containing 75% of the least optimal agents and explores using RT-ARP. |

**Table 4.** Summary of labels for the volatile system

| Label | Summary |
|---|---|
| <nodrop> | ATA-RIA when the system has no network instability. |
| <drop> | ATA-RIA when 1% of agents leave/rejoin the system each episode between episodes 25 and 75. |
| <nosaskr> | ATA-RIA when 1% of agents leave/rejoin the system each episode between episodes 25 and 75 but the RT-ARP and SAS-KR algorithms are disabled. |

Results for each system are show in Figures 4, 5, 6, and 7. Values are shown for the percentage increase or decrease in system utility with the given algorithms in comparison to the baselines described. In the stable system, the baseline is the <optimal> algorithm, in the exploration system, the <rtrap$^0$> algorithm, the volatile system, the <nodrop> algorithm, <large-optimal> for the large system. 75$^{\text{th}}$ percentile bands over the 100 repetitions of each simulation run are shown for Figure 4.

---

[5]For a larger range of tasks, and a greater number of agents in the system, larger values of $(m \times n)$ may be preferable to allow each agent to use trends over longer-term timescales as it will have a greater range of actions it can take.

**Table 5.** Summary of labels for the large system

| Label | Summary |
|---|---|
| `<large-optimal>` | `ATA-RIA` with 10 agents, configured to give the most optimal possible `RT-ARP` performance in the given system. |
| `<large-25>` | `ATA-RIA` in a system of 25 agents |
| `<large-50>` | `ATA-RIA` in a system of 50 agents |
| `<large-100>` | `ATA-RIA` in a system of 100 agents |

A summary of results are shown in Appendix B in Tables 9, 10, 11, and 12 for the stable, exploration, volatile, and large systems respectively. Statistics for each comparison algorithms' utility values are also shown in Appendix B, Table 13. The p-values showing the statistical significance of the system utility values for each simulation datasets' final episode are shown in Table 14[6].

## 6.2 Analysis and discussion

We now look in detail at our simulation results for each system and analyse the behaviours seen.

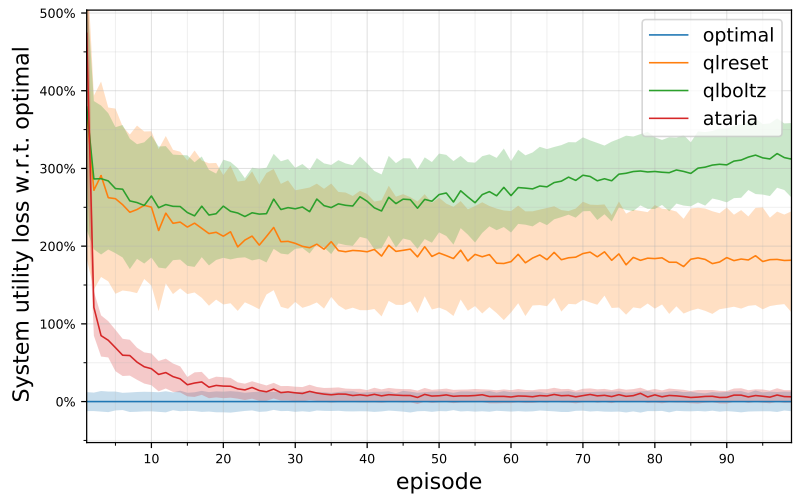

**Fig. 4.** System utility comparison to the system optimal in the stable system

*6.2.1 Stable system.* As seen in Figure 4, the `<ataria>` algorithm performs to 6.7% of the `<optimal>` algorithm after 100 episodes in the *stable system*. Initially $\sim 30\%$ of the atomic task allocations made by the parent agents are not successful, but the failure rate rapidly falls to $< 2\%$. Although exploration is reduced as the algorithm approaches the optimal task allocation strategy, it never fully exploits the best strategy due to the effect of `RT-ARP`, which generates a low level of non-optimal actions. The `<qlreset>` performs 1.8x worse than `<optimal>`. Since values in agents'

---

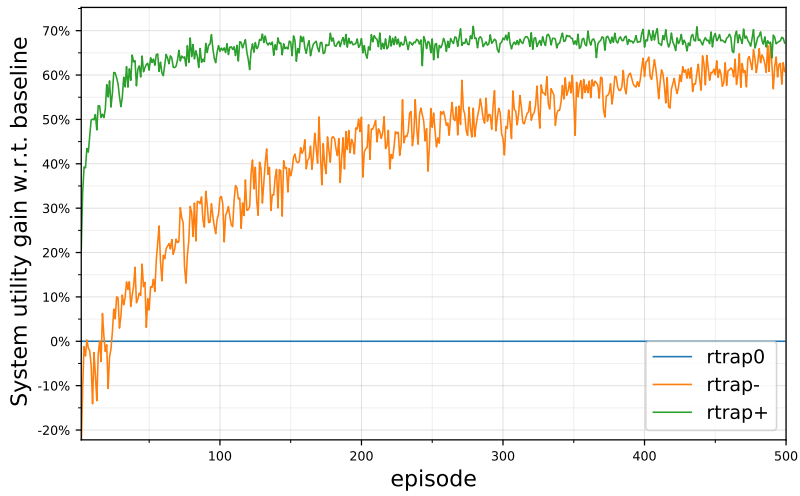

**Fig. 5.** System utility comparison to the system optimal in the exploration system

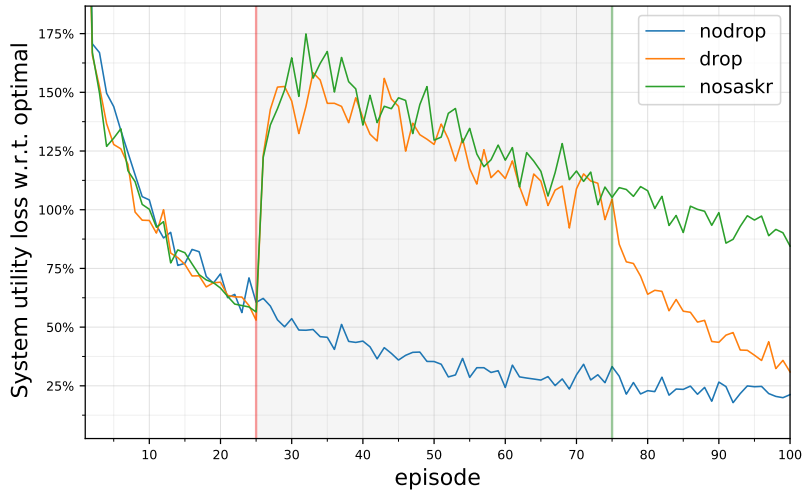

**Fig. 6.** System utility comparison to the system optimal in the volatile system

Q-tables are partially reset each episode, the algorithm fails to use knowledge from past experiences optimally while adapting its policy. Initially the `<qlboltz>` algorithm behaves similarly to `<qlreset>`. However, it explores and learns a policy early in the systems' lifetime. As the system ages, the algorithms' exploration reduces, and it becomes stuck choosing actions based on the initial policy, rather than adapting to changes in the system. As a result, it reaches 2.3x of optimal performance by episode 25, but then worsens to 3.0x by episode 100 as the difference between its stationary policy and the newer, more optimal one increases.

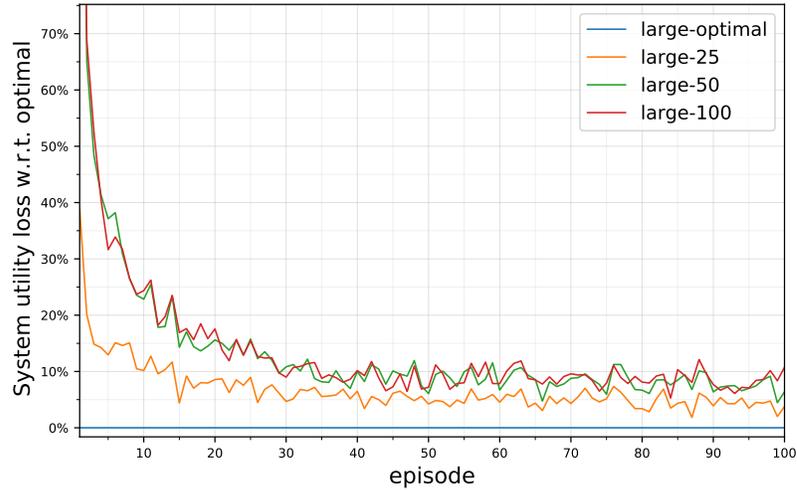

**Fig. 7.** System utility comparison to the system optimal in the large system

Overall, these results show that the `<ataria>` algorithm can optimise system utility well in a stable system. Although the effect of `RT-ARP` means that ATA-RIA is not fully optimal under these conditions, it also improves its ability to adapt to changes as the environment becomes more dynamic.

*Limitations of comparisons based on simulation time.* The average time taken for each algorithm to complete an episode in the stable system as the parent agent count increases is shown in Table 6. As the *large system* involves more parent agents, and the *volatile system* is increasingly non-stationary, the episode times involved for the comparison algorithms proved intractable for useful simulation runs. Due to this, and as the `ATA-RIA` showed better performance in the *stable system*, we only simulate the ATA-RIA algorithm for the systems that follow.

**Table 6.** Runtimes of algorithms in the stable system [7].

| Algorithm | Average time per-episode by parent agent count (secs) | | | | |
|---|---|---|---|---|---|
| | **1 agent** | **2 agents** | **3 agents** | **5 agents** | **10 agents** |
| `<ataria>` | 0.7 | 1.6 | 2.9 | 3.7 | 6.4 |
| `<qlboltz>` | 1.1 | 19.6 | 162.9 | 223.7 | 2302.0 |
| `<qlreset>` | 1.2 | 16.7 | 141.8 | 171.5 | 1862.2 |

*6.2.2 Exploration system.* Next we examine the exploration of state-space in the *exploration system*, in Figure 5. The `<rtrap+>` algorithm gains a 67.0% improvement in system utility compared to `<rtrap0>` after 500 episodes. `<rtrap->` improves 62.7% in task completion performance, with the expectation that this would merge with the utility levels of `<rtrap+>` given more episodes. The RT-ARP algorithm acts of a proxy comparison of the current allocation quality for an agent, to the locally-optimal allocation, and system-optimal allocation qualities for that agent. It drives the agent into better neighbourhoods for its task allocations and increases the systems' utility. As the current neighbourhood nears the optimal neighbourhood for that agent and its tasks, the rate of exploration falls.

*6.2.3    Volatile system.* In the *volatile system* in Figure 6 we see the `SAS-KR` algorithms' effect on system resilience and recovery . Before the impact on agent connectivity is introduced at episode 25, the algorithms' performances are equivalent. On introducing instability, the performance of the `<drop>` and `<nosaskr>` algorithms deteriorate by 72.5%, gradually improving to 59.7% over the course of the disruption. After instability stops at episode 75 `<drop>` recovers to 9.7% of the performance of the non-impacted `<nodrop>` algorithm by episode 100, as compared to 54.6% for `<nosaskr>`.

As the `SAS-KR` algorithm retains the most up-to-date, and least uncertain actions and associated Q-values, better information about past actions and neighbourhoods is kept by the agent as compared to with it disabled. When the instability is removed, the quality of knowledge kept by the `<drop>` algorithm is higher than in `<nosaskr>`, allowing a quicker recovery to more optimal neighbourhood formations, and so task-allocation quality and overall system utility.

*Learning under uncertainty and disruption.* In the stable system, this ability of `SAS-KR` algorithm to retain higher quality knowledge helps guide exploration. When the environment is more disrupted however, it has a greater effect as the agents' knowledge changes more rapidly. The `RT-ARP` algorithm increases exploration during the early episodes when there is large uncertainty in which action choices are optimal. This enables the agent to learn quickly, and slow down learning as performance improves. Similarly it will increase exploration during disruptions as the agents' performance in these environments is most likely less rewarding than in the past. If the disruption is transient, the agent can quickly re-apply its' retained knowledge to recover performance. There may be improvements possible in how well these algorithms perform with further research, however, how the quality of knowledge is judged, and how aggressively the `RT-ARP` algorithm moves between exploration and exploitation, is dependant on the desired behaviour of the specific multi-agent system they are applied in.

*6.2.4    Large system.* The *large system* is shown in Figure 7. Here we see the `<large-25>` algorithm perform within 3.6% of the `<large-optimal>` algorithm, the optimal performance possible for the `ATA-RIA` algorithm in the system. The `<large-50>` and `<large-100>` algorithms optimise system utility to within 7.2% and 8.6% of `<large-optimal>` by the completion of 100 episodes. As expected, the system utility of the `ATA-RIA` algorithm is initially poorer with increasing number of agents in the system. On initialisation of the system, there is a greater likelihood of parent agents being in neighbourhoods with agents that have lower than average atomic task qualities available, or where not all atomic tasks in the parent agents' composite task are completable. There is also a larger system space for the algorithm to search. Even so, the `ATA-RIA` algorithm shows good performance in optimising the system utility to under 10% of optimal with a system of 100 agents.

Although there is a more rapid improvement in utility with less agents since the system-space to learn is smaller, further investigation shows that the performance of the three systems converges with increasing episodes. However, due to the compute and storage limitations of running the simulations repeatedly for longer periods, we have limited the comparisons to 100 episodes. Further research on more powerful simulation platforms would be expected to show a similar behaviour and convergence properties for larger systems.

*6.2.5    Summary.* Overall, the evaluation of the algorithms' presented shows that they perform well at task allocation in both stable and unstable environments, as well as scaling to larger systems. The `ATA-RIA` algorithm improved system utility to 6.7% of the optimal in the simulated system. The `RT-ARP` algorithm reduced exploration as the system utility approached optimal, and adapted well in response to disruption. It allowed agents to alter their neighbourhoods from areas of state-action space that would not allow task completion to those where it would be possible. In environments with disrupted connectivity, the retention of learned knowledge through `SAS-KR` allowed for quicker re-optimisation and adaptation of neighbourhoods, over 5× better than when `RT-ARP` and `SAS-KR` were disabled, and there was no adaptive exploration or knowledge retention strategy.

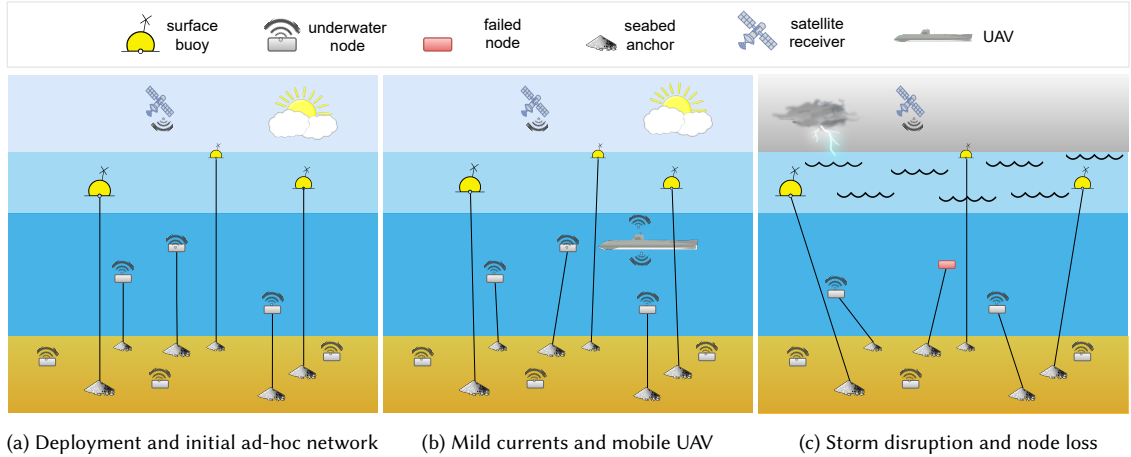

(a) Deployment and initial ad-hoc network    (b) Mild currents and mobile UAV    (c) Storm disruption and node loss

Fig. 8. An illustration of a common Underwater WSN system. In Figure 8a, the nodes are deployed and an initial task optimisation is learned. In Figure 8b, the `ATA-RIA` adapts the actions of nodes to account for movement due to currents, and passing UAVs. In Figure 8c, nodes are highly disrupted and some fail. The `SAS-KR` and `RT-ARP` algorithms work to quickly re-establish an optimal configuration from past knowledge and the prioritisation of exploration as the environment stabilises.

## 7  REAL-WORLD APPLICATIONS

We now detail a real-world system how our work could be applied, relating the behaviour of our algorithms experimentally to the challenges presented. In this paper, we have briefly given examples of realistic Environmental Wireless Sensor Network (EWSN) systems in Examples 3.4, 3.5 and 5.4.

These EWSN examples have been chosen to illustrate the theoretical concepts discussed, the challenges, and key properties of dynamic multi-agent systems. They require ad-hoc learning of agent neighbourhoods. Agents enter and leave the system through component failure and re-deployments. The capabilities of different agents to complete tasks and the qualities they can complete them to can vary due to placement, obstructions, different instrumentation, and wear and tear of components. Our work focuses on adapting to this changeability in the optimal allocation of tasks within a system. However, these examples are also relevant to highlight the types of existing, real-world systems that our work could be applied to, and verified against.

We focus on a sub-category of these systems that contain complex deployments of sensor nodes for ocean monitoring, often described as the Internet of Underwater Things, or the Ocean of Things [7, 31, 32, 56, 91]. The networks built for these systems are termed Underwater Wireless Sensor Networks (UWSN) [11, 22].

### 7.1 Challenges in underwater wireless sensor networks

Figure 8 shows a common UWSN scenario [30, 43, 51, 53, 72] where the deployed nodes can be tethered buoys or submerged sensors, as well as mobile Unmanned Autonomous Vehicles (UAV). In the system shown, all the nodes form ad-hoc communication groups to carry out ocean monitoring tasks such as temperature and salinity measurement [55]. Since radio transmissions are absorbed quickly underwater, acoustic transmission is used. This change affects key properties of the UWSN [30, 45] that our multi-agent algorithms must adapt to if they are to remain useful.

- Acoustic signals have a much lower bandwidth, 10kbps compared to 250kbps for radio transmission.
- Signals propagate over 100,000 times slower than radio.
- Signals travel much further, 10km compared to 100m for radio.
- The nodes' transmission components are hard to recharge, so good power efficiency is essential.
- Underwater links are more unreliable due to corrosion, variable salinity density, and absorption of signals through water. Bit rate errors can be high, and connectivity intermittent.
- Nodes can move under currents, often severely during rough seas, affecting the optimal network configuration.

### 7.2 Algorithm behaviour and impact

*Initial deployment.* The nodes need to form an ad-hoc network to receive task requests and return monitoring data from their sensors (Figure 8a). The `ATA-RIA` algorithm enables nodes to discover each other, establish the capabilities of other nodes, and form groups to optimise their ocean monitoring tasks. Energy resource usage can form part of the quality metric of task completion, encouraging the agents to learn better policies for power efficiency. `RT-ARP` will ensure that they prioritise discovery at this initial stage, and move towards the efficient completion of temperature and salinity sensing tasks as nodes learn more about the system. In our simulations this is the behaviour shown in Figure 4 up to approximately episode 25.

*Behaviour in calm weather.* In stable conditions there will still be some movement of nodes due to currents, replacement of buoys and sensors due to wear, and intermittent mobile nodes such as UAVs (Figure 8b). The `RT-ARP` algorithm maintains a small amount of exploration to not only optimise communication between nodes that have established a neighbourhood group, but also cautiously discover other nodes to add to the group. The `SAS-KR` algorithm helps with transient impacts such as high salinity corrupting node-to-node communications, or UAVs moving in and out of the system. It does so by allowing nodes to retain selective knowledge between last known sightings, so that when these nodes or UAVs reappear, agents can recall that knowledge and quickly re-adjust. In Figure 6 in Section 6 we can see how the `RT-ARP` and `SAS-KR` algorithms achieve this. They help adapt to the intermittent loss of nodes during the period between episodes 25 and 75 by continuing to optimise agents' task completion during that period.

The `N-Prune` algorithm makes sure that throughout the discovery process and adaptation, the limited resources of each node are not overstretched, and the least useful known other nodes are removed from each nodes memory to preserve resource constraints.

*Behaviour in rough seas.* During instability, the effects seen in calm weather are magnified (Figure 8c). In this situation, current reward trends are likely less favourable than those in the past, during calmer conditions. This pushes the `RT-ARP` algorithm towards more extreme exploration as nodes are lost, destroyed, or displaced. As the storm passes and the position of nodes, currents, and salt density stabilises, `SAS-KR` allows the nodes to remember previous nodes it may have lost contact with, or whose capabilities had been disrupted. As `RT-ARP` will accelerate exploration until the systems' performance is comparable to historical values, the use of existing knowledge increase the speed of this recovery by removing the need for nodes to re-learn everything they learned about other nodes prior to disruption. For example, the period of the storm would be similar to the episodes $25 - 75$ in Figure 6 in our simulation, with the ocean calming after 75 episodes. The behaviour of the `RT-ARP` and `SAS-KR` algorithms in accelerating recovery in the simulation should be equivalent to our UWSN system after a disruptive storm.

## 8 CONCLUSIONS

As we have shown in this paper, with the `ATA-RIA` algorithm optimising agents' task allocations, `RT-ARP` adapting exploration based on reward trends, and the `SAS-KR` and `N-Prune` algorithms managing knowledge and neighbourhood retention respectively, the contributions presented here combine to give a novel method of optimising task-allocation in multi-agent systems. The evaluation results show that the combined algorithms give good task allocation performance compared to the theoretical optimal available in the simulated systems, and are resilient to system change with constrained computational cost and other resource usage. This indicates a good basis for successful application to real-life systems where there are resource constraints, and dynamic environments.

The algorithms described here are applicable to a general class of problems where there are dynamic, self-organising networks, and where multiple agents need to learn to associate other agents with subtasks necessary for completion of a composite task. This work may be especially applicable to systems where there are changeable conditions that cause instabilities and where there are very limited possibilities for maintenance or human intervention. There are applications in wireless sensor networks (WSN) [59, 95] where adaptive networking and optimisation are essential to keep usage and maintenance costs minimal. The algorithms' adaptability to connectivity disruption and agent loss indicates that their performance in harsh environmental conditions, and where reliability of components deteriorates over time, may be worth further investigation. Similarly dynamic multi-agent systems such as vehicular ad-hoc networks (VANET) [93], and cloud computing service composition [37, 74], also provide real-world task allocation applications.

Adaptation to congestion when multiple agents are in competition showed how the algorithms could be useful in environments where resource contention on both targets of requests and the network itself are factors. Agents learned to compromise on allocating subtasks to the agents that would give the best quality, but had more competition from other agents, with allocating to agents that had reduced contention on their resources. While this allows a degree of balance to develop in a contained system it would be worth investigating how this behaviour could be used to drive exploration of the greater system. For example, agents who find themselves in a heavily resource competitive area of the system could be pushed to prioritise exploration of less busy areas, adapting their behaviour to not require or utilise the same resources by adopting a different role in the system. This has uses in load balancing workloads across cloud compute systems and energy consumption management in distributed sensor networks.

## A  PARAMETERS FOR SYSTEM SIMULATIONS AND ALGORITHMS

**Table 7.**  General parameter values

| Variable | Summary | Value |
|---|---|---|
| $|AP|$ | Number of atomic task types | 20 |
| $|CP|$ | Number of composite task types | 10 |
| $|K(g)|$ | Size of agents' knowledge | 7 |
| $|N(g)|$ | Size of agents' neighbourhoods | 5 |
| $|ap \in cp|$ | Number of atomic tasks composing a composite task type | 5 |
| n/a | Frequency distribution of composite tasks' arrival in the system | One $cp$ per parent agent per episode |
| $\omega_g$ | The atomic task quality produced by a child agent for a task. | $(0, 1]$ |

**Table 8.**  Simulation parameter values

| Variable | Summary | Optimal | Exploration | Volatile | Large |
|---|---|---|---|---|---|
| $|PG|$ | Number of parent agents in the system | 3 | 3 | 3 | 10 |
| $|CG|$ | Number of child agent in the system | 10 | 10 | 10 | $\{10, 50, 100\}$ |
| $W$ | The approximate action-impact values | {(LINK, 0.10), (INFO, 0.20)} | {(LINK, 0.10), (INFO, 0.20)} | {(LINK, 0.10), (INFO, 0.20)} | {(LINK, 0.10), (INFO, 0.20)} {(LINK, 0.10), (INFO, 0.55)} {(LINK, 0.10), (INFO, 0.60)} |
| $P(leave/join|pg)$ | Probability of agent leaving or re-joining the system each episode | 0 | 0 | 0.01 | 0 |

## B  SUMMARY OF RESULTS

**Table 9.**  Experimental results for the stable system after 100 episodes

| Algorithm | % performance decrease from `<optimal>` (best) |
|---|---|
| `<ataria>` | 6.7% |
| `<qlreset>` | 181.0% |
| `<qlboltz>` | 306.6%(235.0%) |

**Table 10.** Experimental results for the exploration system after 100 episodes

| Algorithm | % performance increase over `<rtrap`$^0$`>` |
|---|---|
| `<rtrap`$^-$`>` | 44.3% |
| `<rtrap`$^+$`>` | 67.0% |

**Table 11.** Experimental results for volatile system after 100 episodes

| Algorithm | % performance decrease from `<nodrop>` |
|---|---|
| `<drop>` | 9.7% |
| `<nosaskr>` | 54.6% |

**Table 12.** Experimental results for large system after 100 episodes

| Algorithm | % performance decrease from `<large-optimal>` |
|---|---|
| `<large-25>` | 3.6% |
| `<large-50>` | 7.2% |
| `<large-100>` | 8.6% |

**Table 13.** Statistics of baseline algorithms results

| Statistic | `<optimal>` | `<rtrap`$^0$`>` | `<nodrop>` | `<large-optimal>` |
|---|---|---|---|---|
| mean | 53.75 | 75.91 | 108.46 | 28.36 |
| std | 0.54 | 8.98 | 22.85 | 6.55 |
| min | 52.55 | 67.75 | 70.30 | 24.93 |
| 25% | 53.38 | 70.31 | 89.47 | 25.63 |
| 50% | 53.80 | 72.12 | 111.61 | 25.98 |
| 75% | 54.12 | 77.76 | 124.32 | 27.72 |
| max | 55.01 | 113.94 | 214.75 | 72.81 |

**Table 14.** Final episode p-values of algorithm results[*].

| Label | p-value | Label | p-value | Label | p-value |
|---|---|---|---|---|---|
| `<optimal>` | 0.54 | `<nodrop>` | 0.86 | `<large-optimal>` | 0.76 |
| `<ataria>` | 0.87 | `<drop>` | 0.87 | `<large-25>` | 0.67 |
| `<rtrap`$^0$`>` | 0.33 | `<nosaskr>` | 0.47 | `<large-50>` | 0.87 |
| `<rtrap`$^-$`>` | 0.29 | | | `<large-100>` | 0.91 |
| `<rtrap`$^+$`>` | 0.50 | | | | |

[*]T-test for the null hypothesis that the utilities in the final episode are equal to the population mean with significance level $\alpha = 0.05$.

## C  CALCULATING APPROXIMATE ACTION-IMPACT VALUES

We ignore actions apart from *INFO* and *LINK* as these are the only ones that alter the neighbourhood, $N$, or knowledge, $K$, of an agent. To make our first approximations we assume that the selection of actions of these two types is distributed uniformly. Given this, the probabilities of changing the neighbourhood or knowledge for these action types is,

$$\text{LINK: } P(N^{'}, N^{''}) = \frac{1}{2} \times \frac{|N|}{|K|}, \text{ and } P(K^{'}, K^{''}) = 0 \tag{27}$$

$$\text{INFO: } P(N^{'}, N^{''}) = 0 \text{ and } P(K^{'}, K^{''}) = 1 - P(N^{'}, N^{''}) \tag{28}$$

For *LINK* actions, an agent will be replaced in $N$ with one of the agents in $K$. We make the simplification that the neighbourhood impact of the actions will be 1 if we add an agent that can complete a task type better than an existing agent in the neighbourhood, and 0 otherwise. Assuming agents are distributed randomly across $N \cup K$, on average a *LINK* action will produce a neighbourhood impact of $1 - \frac{N}{K}$. Therefore we get an approximation of action impact of,

$$\left( \frac{1}{2} \frac{|N|}{|K|} \right) \left( 1 - \frac{|N|}{|K|} \right) \tag{29}$$

For *INFO* actions an agent in $K$ will be replaced by one in $G$. Again, assuming agents are distributed randomly across $K \cup G$, on average an *INFO* action will produce a knowledge impact of $1 - \frac{|K|}{|G|}$. Therefore,

$$\left( 1 - \frac{1}{2} \frac{|N|}{|K|} \right) \left( 1 - \frac{|K|}{|G|} \right) \tag{30}$$

So that,

$$W = \left\{ \left( \text{LINK}, \left[ \frac{1}{2} \frac{|N|}{|K|} \right] \left[ 1 - \frac{|N|}{|K|} \right] \right), \left( \text{INFO}, \left[ 1 - \frac{1}{2} \frac{|N|}{|K|} \right] \left[ 1 - \frac{|K|}{|G|} \right] \right) \right\} \tag{31}$$

*Example C.1 (Optimal allocations in multi-agent systems).* A system contains 100 agents, with each agents' neighbourhood size being 10, and its knowledge 20. Using the above approximation we get action-impact values,

$$W = \left\{ \left( \text{LINK}, \frac{1}{8} \right), \left( \text{INFO}, \frac{3}{5} \right) \right\} \tag{32}$$